\documentclass[11pt]{article}

\usepackage[preprint]{acl}

\usepackage{times}
\usepackage{latexsym}

\usepackage[T1]{fontenc}
\usepackage[utf8]{inputenc}

\usepackage{microtype}

\usepackage{inconsolata}

\usepackage{graphicx}

\usepackage{amsmath,amsfonts,bm}

\def\eqref#1{equation~\ref{#1}}
\def\1{\bm{1}}

\DeclareMathAlphabet{\mathsfit}{\encodingdefault}{\sfdefault}{m}{sl}
\SetMathAlphabet{\mathsfit}{bold}{\encodingdefault}{\sfdefault}{bx}{n}

\usepackage{hyperref}
\usepackage{url}

\usepackage{booktabs}       % professional-quality tables
\usepackage{amsfonts}       % blackboard math symbols
\usepackage{nicefrac}       % compact symbols For 1/2, etc.
\usepackage{microtype}      % microtypography
\usepackage{xcolor}         % colors
\usepackage{threeparttable}

\usepackage{xspace}

\usepackage{colortbl}
\usepackage{enumitem}
\usepackage{makecell}
\definecolor{colorTab}{rgb}{0.9,0.9,0.98}
\definecolor{color3}{gray}{0.95}

\definecolor{css}{rgb}{0.7529, 0, 0}
\definecolor{fss}{rgb}{0, 0.7, 0.3}
\definecolor{pbp}{rgb}{0.2, 0.2, 0.6}

\usepackage{amsmath}
\usepackage{amsthm}
\usepackage{amsfonts}
\usepackage{amssymb}
\usepackage{mathtools}
\usepackage{nicefrac}
\usepackage{algorithm}
\usepackage{multicol}
\usepackage{algorithmicx}
\usepackage{algpseudocode}

\usepackage{graphicx}
\usepackage{subcaption}  % 替代 subfigure
\usepackage{wrapfig}
\usepackage{booktabs}
\usepackage{array}
\usepackage{multirow}
\usepackage{adjustbox}
\usepackage{lscape}
\usepackage{caption}
\usepackage{wrapfig}
\usepackage{arydshln}
\usepackage[table]{xcolor}
\definecolor{codeblue}{rgb}{0.25, 0.5, 0.5}
\definecolor{codekw}{rgb}{0.35, 0.35, 0.75}
\definecolor{Gray}{gray}{0.95}

\usepackage{pifont}
\usepackage{bbding}

\usepackage{listings}
\lstdefinestyle{Pytorch}{
    language         = Python,
    backgroundcolor  = \color{white},
    basicstyle       = \fontsize{8.0pt}{9pt}\selectfont\ttfamily\bfseries,
    columns          = fullflexible,
    breaklines       = true,
    captionpos       = b,
    commentstyle     = \fontsize{4pt}{4pt}\color{codeblue},
    keywordstyle     = \fontsize{4pt}{4pt}\color{codekw},
    morekeywords     = {with,scatter_,norm,sort},
}

\theoremstyle{plain}

\theoremstyle{definition}

\theoremstyle{remark}

\newlength\savewidth

\def\eg{\emph{e.g.}}

\title{BTC-LLM: Efficient Sub-1-Bit LLM Quantization via Learnable Transformation and Binary Codebook}

\author{
\textbf{Hao Gu$^{1}$\thanks{Equal contribution.}}
\quad
\textbf{Lujun Li$^{1}$\footnotemark[1]}
\quad
\textbf{Hao Wang$^{2}$}
\quad
\textbf{LEI WANG$^{1}$}
\quad
\textbf{Zheyu Wang$^{3}$}
\\
\textbf{Bei Liu$^{1}$}
\quad
\textbf{Jiacheng Liu$^{1}$}
\quad
\textbf{Qiyuan Zhu$^{1}$}
\quad
\textbf{Sirui Han$^{1}$\thanks{Corresponding authors.}}
\quad
\textbf{Yike Guo$^{1}$\footnotemark[2]}
\\[0.5em]
$^{1}$The Hong Kong University of Science and Technology
\qquad
$^{2}$City University of Hong Kong
\\
$^{3}$University of Electronic Science and Technology of China
\\
\texttt{marcusguhao@gmail.com}
\qquad
\texttt{siruihan@ust.hk}
\qquad
\texttt{yikeguo@ust.hk}
}

\begin{document}
\maketitle
\begin{abstract}
Binary quantization represents the most extreme form of compression, reducing weights to $\pm$1 for maximal memory and computational efficiency. While recent sparsity-aware binarization achieves sub-1-bit compression via weight pruning, it faces critical challenger: performance degradation, mask-management overhead, and limited hardware compatibility. In this paper, we present BTC-LLM, a novel sub-1-bit LLM quantization framework that leverages binary pattern clustering and weight transformation to overcome these limitations. Our approach incorporates two key innovations: 
(1) a Binary Codebook that clusters recurring vectors into compact indices using custom distance metrics and sign-based updates; (2) a Learnable Transformation that reduces outliers and promotes shared sign patterns among binary weights.
This eliminates sparse masks, enabling efficient inference on standard hardware. Extensive evaluations across LLaMA, Qwen, and FBI-LLM families demonstrate that BTC-LLM achieves state-of-the-art results in extreme compression (1.11–0.7 bits). Notably, BTC-LLM compressed to 0.8 bits on LLaMA-2-13B maintains high performance—with only a 3.1\% accuracy drop in zero-shot benchmarks—while delivering a 1.6$\times$ speedup over FP16.

\end{abstract}

\section{Introduction}

\begin{figure}
    \centering
    \includegraphics[width=1\linewidth]{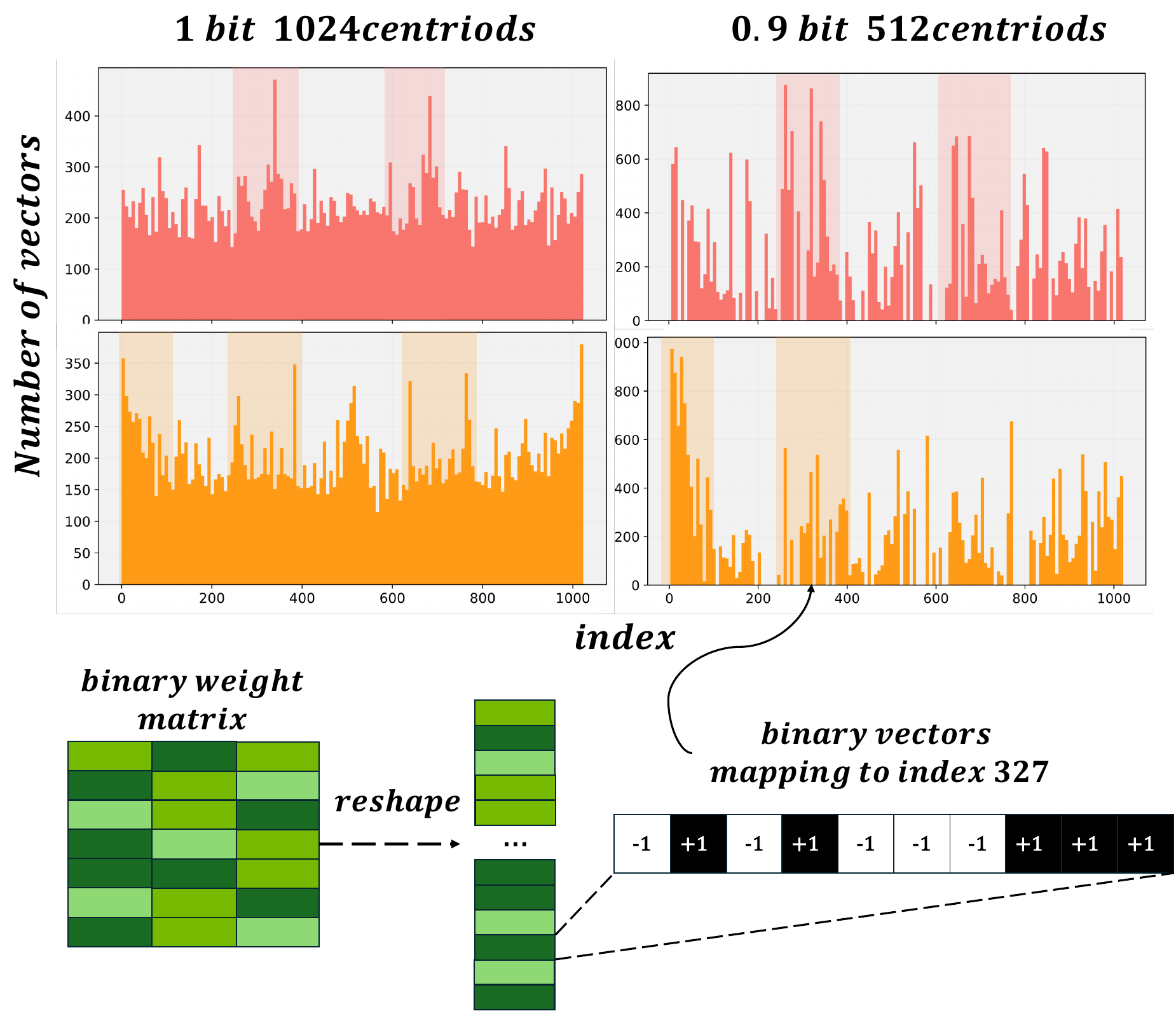}
    \caption{Binary vector distribution (length 10). \textbf{Left}: Standard mapping to 1024 indices. \textbf{Right}: 512 codebook centroids.}
    \vspace{-5mm}
\label{fig:dis}
\end{figure}
Recent Large Language Models (LLMs) such as GPT-4o~\citep{gpt4o} and DeepSeek-R1~\citep{guo2025deepseek} have revolutionized natural language processing (NLP), achieving state-of-the-art performance across diverse tasks~\citep{wei2022emergent}. However, the massive scale of models like DeepSeek-R1 (671B parameters) creates unsustainable memory and storage requirements, preventing practical deployment in constrained environments. Model quantization~\citep{ma2024affinequant} addresses this by reducing numerical precision, slashing memory usage by 4$\sim$8× with minimal accuracy drop. Recent advances, such as Omniquant~\citep{shao2023omniquant} and DuQuant~\citep{Lin2024DuQuantDO}, demonstrate that even sub-4-bit methods can maintain > 90\% of original model performance.

\begin{figure*}
    \vspace{-3mm}
    \centering
    \includegraphics[width=1\linewidth]{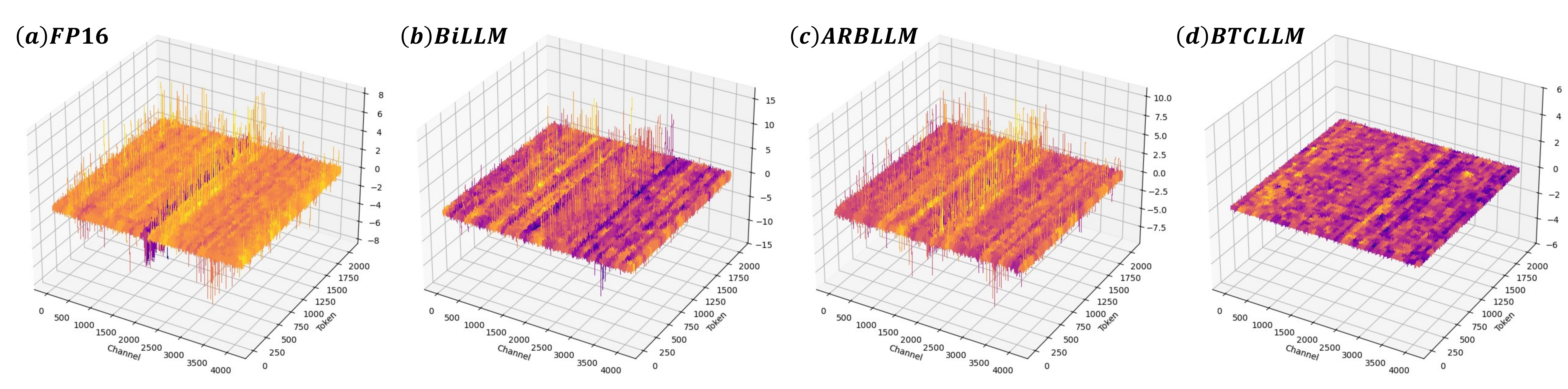}
    \caption{Activation distributions for the \texttt{self\_attn.k\_proj} layer in the LLaMA-2-7B model: (a) Original FP16 (max abs:8), (b) BiLLM (max abs:15), (c) ARB-LLM (max abs:10), and (d) our proposed BTC-LLM (max abs:0.4).}
    \label{fig:weight_dist}
    \vspace{-3mm}
\end{figure*}
Binary quantization~\citep{ref07_xnor} represents the most aggressive quantization approach, converting floating-point weights to binary values ($\pm1$) to reduce memory requirements by over 32$\times$~\citep{liu2018bi}. For instance, BitNet~\citep{wang2023bitnet} pioneered QAT for 1-bit LLMs, achieving low memory consumption (0.4GB) and fast inference (29ms). PTQ methods like BiLLM~\citep{Huang2024BiLLMPT} and ARB-LLM~\citep{li2024arb} employ advanced binarization strategies (\eg, residual approximation, alternating refinement) to enhance 1-bit LLM performance without retraining. STBLLM~\citep{dong2024stbllm} removes redundant binary parameters to achieve sub-1-bit compression with semi-structured N:M sparsity. However, such sparsity-based binarization faces critical challenges:
\textbf{(1) Performance Collapse:} STBLLM relies on detecting which elements to prune, yet suffers from accuracy degradation, retaining only 51$\sim$65\% of full-precision performance on challenging benchmarks (\eg, ARC-c and HellaSwag).
\textbf{(2) Hardware Incompatibility:} Structured sparsity such as $2{:}4$ is not a free lunch. In a 4-value tuple, the $2{:}4$ pattern admits ${4 \choose 2}=6$ possible mask configurations, requiring {$\lceil \log_2 6 \rceil = 3$} bits to encode. Consequently, the effective storage cost per weight is 

$$\frac{\text{sign bits(2)} + \text{mask bits(3)}}{\text{number of weights(4)}} = 1.25 \text{ bits/weight}.$$
These naturally yield a question: \textbf{(RQ) How can we design a hardware-friendly algorithm to further compress binary weights for sub-1-bit LLMs while maintaining performance?} To address this, we first analyze the weight distribution patterns of binarized LLMs to explore potential for more compact compression. As shown in {Figure~\ref{fig:dis}}, we adopt product quantization by splitting the binary weight matrix into sub-vectors, each mapped to an index (e.g., index 327 corresponds to the binary pattern [-1, +1, -1, +1, ...]). Interestingly, these locally continuous blocks exhibit clear clustering patterns, which motivates us to further compress the model by representing redundant $\pm1$ weights with a compact set of centroid vectors.

We further examine the activation distribution of binarized LLMs and empirically observe the presence of prominent outliers. Such large activations amplify the quantization error, since the forward error term can be expressed as \texttt{$XW-X\widehat{W}=X(W-\widehat{W}),$} where outlier entries in $X$ magnify the impact of binarized weight noise. As shown in Figure~\ref{fig:weight_dist}(b-c), BiLLM shows a wide dynamic range (with absolute values up to 15) with prominent outliers, while ARB-LLM still exhibits noticeable noise and instability. This motivates the need for outlier mitigation, even in binarization methods.

Building on these insights, we propose BTC-LLM, a novel framework that enables extreme compression of LLMs to below 1 bit per parameter. Our approach adopts a two-pronged strategy to tackle key challenges.
\textbf{First, we propose the Flash and Accurate Binary Codebook to leverage binary weight redundancy.} This hardware-friendly approach achieves sub-1-bit compression without the overhead of sparse masks, maintaining model performance by preserving key distributional characteristics.
\textbf{Second, to mitigate activation outliers and encourage cluster in binary weights, we introduce a Learnable Transformation} consisting of an invertible parameter $D_{\pm}$ and $P$.
Fig.~\ref{fig:weight_dist}(d) shows that this approach suppresses activation outliers, capping the maximum magnitude at 0.4.

\begin{figure}
    \centering
    \includegraphics[width=0.9\linewidth]{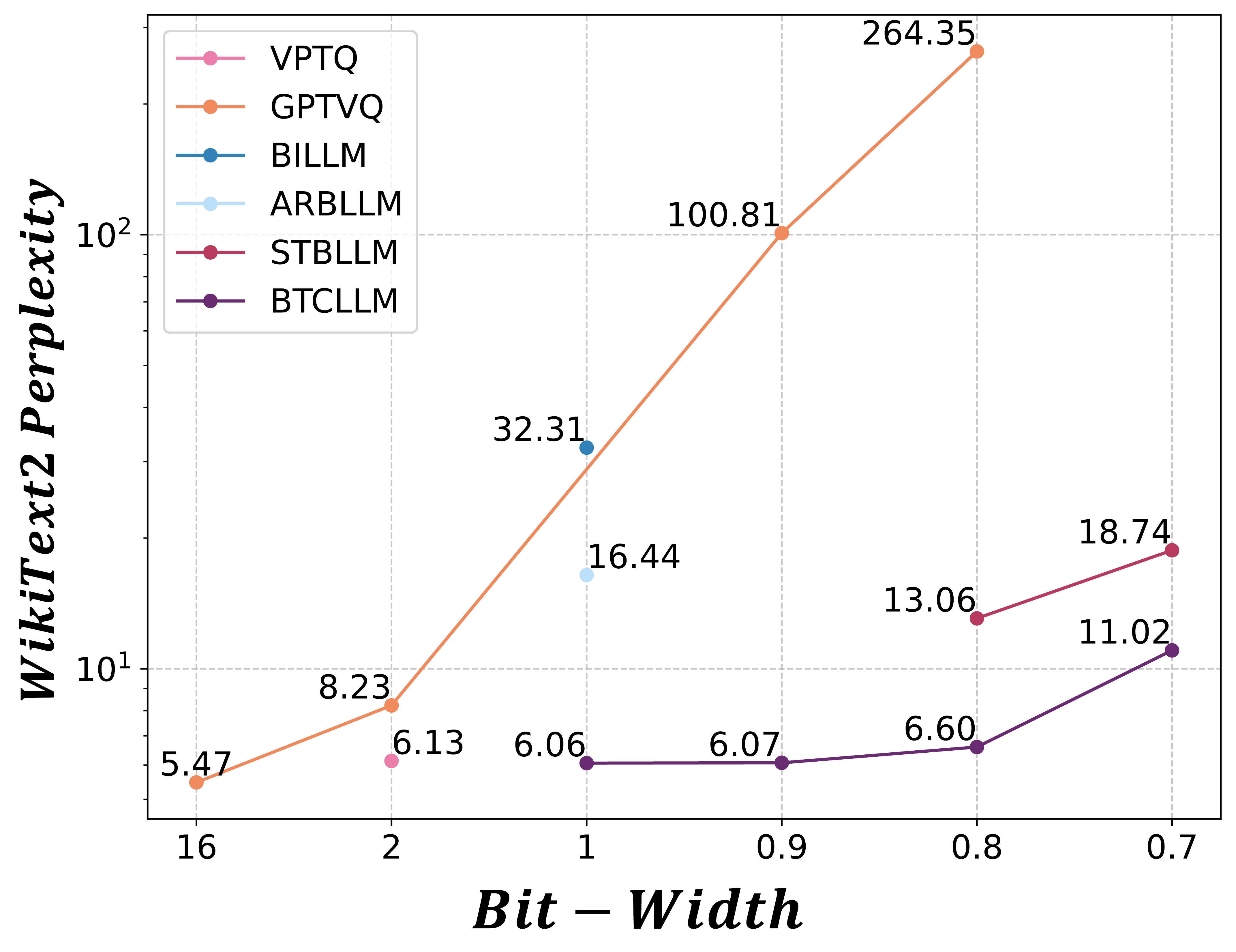}
    \caption{Perplexity of  LLaMA-2-7B on WikiText2. Our BTC-LLM outperforms 2-bit methods at 0.9-bit.}
    \label{fig:ppl} 
    \vspace{-2mm}
\end{figure}
As shown in Figure~\ref{fig:ppl}, our comprehensive evaluations of the LLaMA family of models (7B to 65B parameters) demonstrate the performance of BTC-LLM in multiple bit width settings. Our binary baseline achieves 6.06 PPL, outperforming even 2-bit quantization methods. BTC-LLM remains robust at 0.9–0.8 bits, matching its 1.11-bit performance. Even at 0.7 bits, it achieves 11.02 PPL with 22$\times$ memory savings. It significantly surpasses STBLLM in zero-shot tasks (e.g., +5.0\% on LLaMA-2-13B at 0.8 bits), proving its superior efficiency under extreme quantization.

\section{Related Work}
\noindent\textbf{LLM Quantization} reduces memory and computation by representing parameters with fewer bits.
The pioneering Quantization-aware Training (QAT) methods like LLM-QAT~\citep{liu2023llm} can achieve excellent results but require extensive retraining that is expensive for billion-parameter LLMs.
Existing PTQ methods fall into two main categories: (1) scaling-based approaches, such as AWQ~\citep{lin2024awq} and SmoothQuant~\citep{xiao2023smoothquant}, which identify and rescale influential weights to control activation outliers; and (2) rotation-based approaches, such as QuIP\#~\citep{tseng2024quip} and QuaRot~\citep{ashkboos2024quarot}, which redistribute outliers more evenly across dimensions with transformations.

\noindent\textbf{Binarization} represents the most extreme form of quantization, constraining parameters to a single bit ($\pm 1$). It was first explored in CNNs with XNOR-Net~\citep{rastegari2016xnor} and Bi-Real Net~\citep{liu2018bi}, and later extended to LLMs by BitNet~\citep{wang2023bitnet}, which showed the feasibility of training 1-bit models from scratch. Recent PTQ methods for LLMs include BiLLM~\citep{huang2024billm}, which preserves salient weights, and ARB-LLM~\citep{li2024arb}, which iteratively refines bias and scaling factors. To push beyond 1 bit, STBLLM~\citep{dong2024stbllm} introduced sparsity on binary weights for sub-1-bit compression.

\section{Preliminary}
\paragraph{Binarization.} Binarization represents an extreme form of weight compression in LLMs. For a full-precision weight $\mathbf{W}\in \mathbb{R}^{ n \times m}$, we define the objective of binarization as
\begin{align}
&\mathop{\arg\min}\limits_{\alpha,\mathbf{B}}||\widetilde{\mathbf{W}}-\alpha \mathbf{B}||^2_F,
\\
\widetilde{\mathbf{W}} &= \mathbf{W} -  \mu,\,\mu= \frac{1}{m}\sum_{j=1}^m\mathbf{W}_{.j}\, , 
\label{eq:binary}
\end{align}
where $\alpha \in \mathbb{R}^n$ denotes the row-wise scaling factor, and $\mathbf{B} \in \{+1, -1\}^{ n \times m}$ is a binary matrix. 

It is a common practice to apply a row-wise redistribution before binarization first to achieve a zero-mean distribution in a row. Under the objective of binarization (Equation~\ref{eq:binary}), the optimal solutions for $\alpha$ and $\mathbf{B}$ can be solved with $\alpha=\frac{1}{m}\sum_{j=1}^m|\widetilde{\mathbf{W}}_{.j}|$ and $\mathbf{B}=\operatorname{sign}(\widetilde{\mathbf{W}})$ respectively. However, simply applying this strategy can incur substantial $L_1$ binarization error for LLMs, formulated as: \begin{equation} L_{1} = ||R||^2_F,\quad \text{where} \ R = W - \alpha_1 B_1 - \mu, \end{equation}
To mitigate this error, different approaches have been proposed. BiLLM~\citep{huang2024billm} considers salient weights and approximates the residual with a secondary binarization $R \approx \alpha_2 B_2$. In contrast, ARB-LLM~\citep{li2024arb} addresses the distribution shift between the means of binarized and full-precision weights by iteratively refining the bias $\mu_{\text{refine}} = \mu + \frac{1}{m} \sum_{j=1}^{m} R_{\cdot j}$, the row scaling factor $\alpha_{\text{refine}} = \frac{1}{m} \mathrm{diag}(B^\top (W - \mu_{\text{refine}}))$, and the binarized matrix $B_{\text{refine}} = \mathrm{sign}(W - \mu_{\text{refine}})$.

\paragraph{Codebook Compression.} Pruning is appealing in principle, but it often leads to accuracy degradation and non-trivial mask-index overhead. As noted in introduction, semi-structured pruning requires $0.25$ mask bits per weight.
Besides scalar quantization, vector quantize~\citep{liu2024vptq,van2024gptvq} employs a codebook to represent weights. To be specific, a weight $\mathbf{W}_{ n \times m}$ is mapped in a codebook $\mathbf{C}_{ c \times v}$ where $v$ is the sub-vector length and $c$ is the number of centroids. Now we need to store the codebook $\mathbf{C}_{ c \times v}$ as well as the index assignments instead of the original weights. Since the codebook overhead can be ignored and the compression ratio can be calculated as ${\lceil \log_2 c \rceil} / ({16 \cdot v})$ bits of weights index storage.

% \paragraph{Learnable Transformation.} Recent work \cite{xiao2023smoothquant, shao2023omniquant, liu2024spinquant, ashkboos2024quarot, sun2024flatquant, hu2025ostquant} on weights, activations quantization have shifted the focus towards eliminating outliers.
% Outliers enlarge the value range, leading to a coarser quantization step size $\text{scale}=\text{max(value)}/2^n$, which amplifies quantization error. Formally, $XW-X\widehat{W} = X(W-\widehat{W})$, where outlier entries in $X$ magnify the effect of quantized weight noise. However, the outlier issue remains unexplored in the context of binarized LLMs, where quantization noise is inherently more severe.

\section{Methodology}
\begin{figure*}[t]
    % \vspace{-5mm}
    \centering
    \includegraphics[width=1\linewidth]{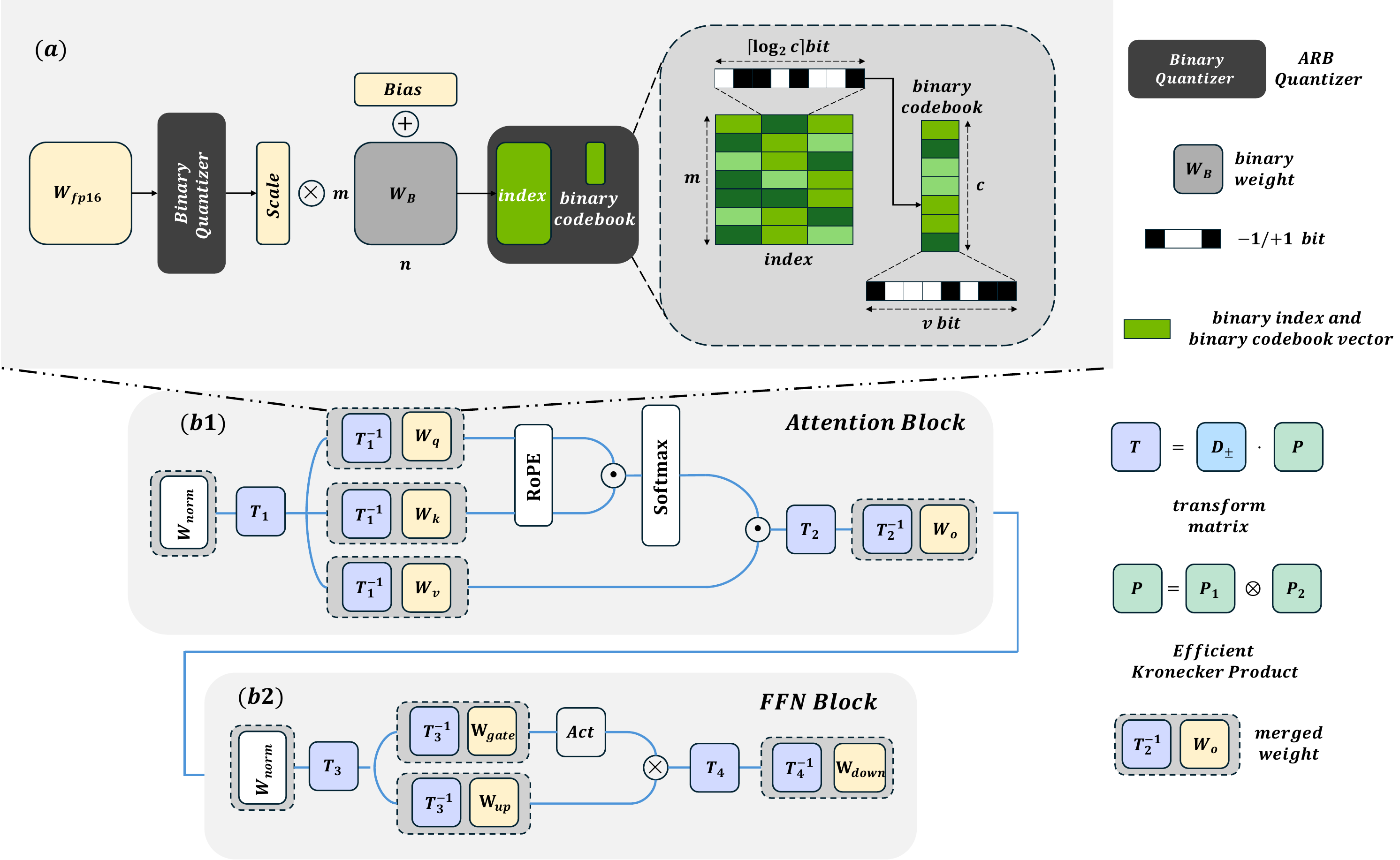}
    \caption{
    Overall architecture of BTC-LLM.
    \textbf{(a)} Sub-bit pipeline: the ARB quantizer transforms full-precision weights into binary form with associated scale and bias, followed by binary codebook representation and index assignment.
    \textbf{(b)} Structure of transformed attention (b1) and FFN (b2) blocks. The transform matrix is merged into the weights to ensure computational equivalence and efficiency.
    }
    \label{fig:method}
    \vspace{-2mm}
\end{figure*}

As shown in Figure~\ref{fig:method}, we introduce BTC-LLM, a novel sub-1-bit LLM quantization method combines a Flash and Accurate Binary Codebook to capture repeated $\pm 1$ patterns with a learnable incoherence-processing transform that reduce outilers and aligns weights to the codebook.

% , and calculating full precision Hessian-weighted distances requires high cost.

\subsection{Flash and Accurate Binary Codebook}
\label{method:binary_codebook}
\textbf{Binary Codebook.} Existing vector quantization methods~\citep{liu2024vptq,van2024gptvq} are tailored for full-precision weights and are misaligned with the nature of binary weights, directly apply a sign function to full-precision codebooks resulting in significant errors. To address this mismatch, we introduce a binary-specific codebook tailored for compressing binarized weights.

Although both the codebook entries and weights are constrained to $-1$ and $+1$, finding the optimal codebook remains an NP-hard problem, detail refer to Appendix~\ref{np-hard}. To solve this, we propose an efficient approximate optimization method inspired by the floating-point KMeans algorithm~\citep{IKOTUN2023178}, combined with the feature of binary vectors distribution. The process consists of three main stages:

\textbf{(1) Initialization}: 
Given binary vectors $\mathbf{B} = \{\mathbf{b}_1, \mathbf{b}_2, \dots, \mathbf{b}_N\}$ where $\mathbf{b}_i \in \{-1, +1\}^v$, we extract the set of unique vectors $\mathcal{U} = \{\mathbf{u}_1, \dots, \mathbf{u}_M\}$ from $\mathbf{B}$. If $M \geq K$ (codebook size), we select the top-$K$ most frequent vectors in $\mathcal{U}$ as the initial centroids $\mathcal{C}^{(0)} = \{\mathbf{c}_1^{(0)}, \dots, \mathbf{c}_K^{(0)}\}$. Otherwise $M \textless K$, we set $\mathcal{C}^{(0)} = \mathcal{U}$ and let $K = M$.

\textbf{(2) E-step Assignment}:  
For each vector $\mathbf{b}_i$, we first test whether it is identical to any centroid $\mathbf{c}_k$; if so, we simply set $z_i = k$. Otherwise, we choose the nearest centroid via $z_i = \operatorname*{arg\,min}_{k} \bigl\lVert \mathbf{b}_i - \mathbf{c}_k \bigr\rVert_2^{2}.$
Because every element is binary ($\pm 1$), the squared Euclidean distance reduces to a Hamming distance:
\begin{align}
\bigl\lVert \mathbf{b} - \mathbf{c} \bigr\rVert_2^{2}
= \sum_j (b_j - c_j)^2 =
\\
4 \sum_j \bigl[ b_j \neq c_j \bigr]
= 4 \, d_H(\mathbf{b}, \mathbf{c}),
\end{align}
where $d_H(\mathbf{b}, \mathbf{c})$ counts the number of different elements. By packing the $\pm 1$ entries into \texttt{int64}, the Hamming distance can be computed with one \texttt{XOR $\to$ POPCNT} instruction: $d_H \left(\mathbf{b}, \mathbf{c}\right)=$ \texttt{POPCNT} $\left(\mathbf{b} \oplus \mathbf{c}\right)$~\citep{tairenpiao_xnor_gemm, pham2025xnor}. Unlike reconstruction error-based metrics such as $\|X \mathbf{B} - X \hat{\mathbf{B}}\|_2^2$, this approach directly leverages the binary structure, avoiding costly matrix multiplications.

\textbf{(3) M-step Centroid Update}:
For cluster $k$ with assignment set $\mathcal{B}_k\subset\{\pm1\}^L$, we update the binary centroid $\mathbf{c}_k\in\{\pm1\}^L$ by solving:
$$
\mathbf{c}_k=\operatorname{sign}\!\Big(\tfrac{1}{|\mathcal{B}_k|}\sum_{\mathbf{b}_i\in\mathcal{B}_k}\mathbf{b}_i\Big),\quad
\operatorname{sign}(0)=+1.
$$
This keeps the centroid still in binary.

After initialization, we alternate \textbf{E–step} and \textbf{M–step}: the E-step assigns each binary vector to its nearest codeword, yielding index $\mathbf{z}$; the M-step updates the codebook $\mathbf{C}$. Both steps are implemented with bit-packing and XNOR/POPCNT, rather than costly floating-point reductions. This full binarization approach bypasses the computationally expensive distance metrics inherent in full-precision K-means, and eliminates the need for complex variants such as Hessian-weighted K-means. 

% To further reduce memory footprint and I/O overhead, we employ a shared codebook across all attention block and all FFN block. During inference, this single codebook is cached and reused, significantly lowering parameter loading requirements and bandwidth pressure. 

To fully exploit the efficiency of binary arithmetic, we propose \textbf{Binary Codebook LUT-GEMM}, following the lookup-table GEMM design in~\citep{guo2024fast,wei2024tmaccpurenaissancetable} to accelerate inference with binary weights. The key idea is to replace most multiply–accumulate operations with table lookups. Since our weights are stored as a binary codebook, only a finite set of binary patterns can appear, and these patterns are repeatedly reused across the matrix. Given an input activation, we first compute the dot products between the activation blocks and all codebook patterns once, and store these results in a small lookup table. Consequently, the GEMM operation is simplified to \textbf{a lookup-and-accumulate process}: we simply fetch the pre-computed results using the weight indices and sum them up. This design avoids any runtime dequant and significantly reduces arithmetic cost by maximally reusing the activation–pattern partial sums. More details refer to appendix~\ref{app:lut}.

\subsection{Learnable Transformation}
\paragraph{Transformation Pair.} To address the outlier issue in binarized LLM and aligns weights to the codebook, in this section, we propose learnable transform scheme to reduce quantization error. 
Specifically, we introduce two learnable parameters, $D_{\pm}$ and $P$, combining them into a transformation pair $T := D_{\pm} P$, 
where $D_{\pm}=\operatorname{diag}(\sigma)$ denote a diagonal sign matrix with $\sigma_i\in\{\pm1\}$, being invertible, it performs channel-wise sign flips without changing magnitudes and $P$ is a learnable invertible affine matrix. Following FlatQuant~\citep{sun2024flatquant}, we parameterize $P$ as a Kronecker product $P=P_1\otimes P_2$ and directly update the lightweight factors $(P_1,P_2)$. Its inverse is computed via $P^{-1}=P_1^{-1}\otimes P_2^{-1}$, enable efficient online transformation.

\paragraph{Optimization detail.} We learn the discrete sign flips $D_{\pm}$ using a straight-through estimator (STE), and apply a larger learning rate to $D_{\pm}$ for stable updates. 
The affine transform $P$ is optimized with standard gradient descent. We optimize the transform parameters in a block-wise manner. For the $l$-th Transformer block, we solve
\begin{equation}
\min_{\mathbf{T}_l} \Big( \big|\big| \mathcal{F}_l(X) - \hat{\mathcal{F}}_l(X;{T}_l) \big|\big|_F^2 + \mathcal{L}_{\text{aux}} \Big),
\end{equation}
where $\mathcal{F}_l(\cdot)$ and $\hat{\mathcal{F}}_l(\cdot)$ denote the original and quantized block, and ${T}_l$ collects the transformation parameters for that block.

To facilitate efficient representation via a compact codebook, we introduce $\mathcal{L}_{\text{aux}}$. By regularizing the Gram matrix, this term encourages binary weights to reuse the same sign patterns (see Figure~\ref{fig:dis}). Specifically, stack the $B$ binary vectors $\{\mathbf{b}_1, \mathbf{b}_2, \dots, \mathbf{b}_B\}$ into a row matrix $M\in\{\pm1\}^{B\times v}$ and define the vector-similarity Gram matrix $G=\tfrac{1}{v}MM^\top\in\mathbb{R}^{B\times B}$. When vectors share similar sign patterns, the energy of $G$ concentrates on its top-$K$ eigenvalues $\{\lambda_i(G)\}_{i=1}^K$. We encourage this by minimizing $\mathcal{L}_{\text{sim}} = \operatorname{Tr}(G) - \sum_{i=1}^{K} \lambda_i(G)$. To avoid the collapse where all entries are $+1$ or $-1$, we add a global balance term that keeps the overall sign mean near zero,
$\mathcal{L}_{\text{bal}} = \Big(\tfrac{1}{Bv}\sum_{b=1}^{B}\sum_{\ell=1}^{v} M_{b,\ell}\Big)^{2},$
and define the auxiliary objective $\mathcal{L}_{\text{aux}} = \lambda_1\,\mathcal{L}_{\text{sim}}+\lambda_2\,\mathcal{L}_{\text{bal}}$.

\paragraph{Integration into Transformer Architecture.}
For each linear layer $Y=XW^\top$, we apply the invertible transform $T$ and reparameterize the weight as
\begin{equation}
Y = XW^\top \equiv (XT)\,(T^{-1}W^\top),
\end{equation}
which leaves the output unchanged in full precision and $T^{-1}$ can be comptute effciently with kernel. We then binarize and compress only the reparameterized weights,
\begin{equation}
\hat W \;=\; \mathbf{Codebook}\big(\mathbf{B}(T^{-1}W^\top)\big),
\end{equation}
where $\mathbf{B}(\cdot)$ is the binary quantizer and $\mathbf{Codebook}(\cdot)$ denotes codebook compression. During inference, we apply the transform $T$ to activations on-the-fly ($X←XT$) and compute the output via LUT-based GEMM, leveraging the index from our binary codebook.

For the binary quantizer, we follow the binarization procedure described in ARB-LLM. Since the incoherence-processing transformation inherently incorporates activation information, we specifically adopt the naive ARB method rather than the ARB-RC or ARB-X variants for weight binarization which is faster and simpler.

\subsection{Compression Analysis}
As illustrated in Figure~\ref{fig:method} (a), binary weights are compressed into a binary codebook and index mappings. Given an original weight matrix of shape $n \times m$, with a codebook of size $c$ and vector length of $v$, the index requires $\lceil \log_2 c \rceil$ bits per vector, and each centroid occupies $v$ bits. In Figure~\ref{fig:method} (b), the transformation matrix can be fused into the model weights, incurring no additional storage overhead. Thus, the total storage cost is $vc + \lceil \log_2 c \rceil \cdot mn / v$. Since $vc$ is relatively small and can be amortized, the effective compression ratio is approximately ${16 \cdot v} / {\lceil \log_2 c \rceil}$.

\begin{table*}[!t]
\vspace{-2mm}
\renewcommand{\arraystretch}{1.0}
\footnotesize
\centering
\caption{Perplexity results comparison on the LLaMA family.\label{tab:llama_series}}
% \vspace{2mm}
\resizebox{1.0\textwidth}{!}
{
\begin{tabular}{lcrrrrrrrrrrr}
    \toprule
    \rowcolor{color3}
    \multicolumn{2}{c}{\textbf{Settings}} & \multicolumn{4}{c}{\textbf{ LLaMA-1}}  & \multicolumn{2}{c}{\textbf{ LLaMA-2}} & \multicolumn{1}{c}{\textbf{ LLaMA-3}}\\
    \midrule
    Method & \multicolumn{1}{p{4em}}{W-Bits} & 7B    & 13B   & 30B   & 65B   & 7B    & 13B   & 8B\\
    \midrule
    FP16   & 16   & 5.68  & 5.09  & 4.1   & 3.53  & 5.47  & 4.88 & 6.14 \\
    \cdashline{1-11}
    \addlinespace[0.2em]

    QuIP\# & 2    & 6.86  & 5.97  & 5.02  & 4.36  & 6.66  & 5.74  &   -  \\
    GPTVQ  & 2.15 & 9.64  & 6.58  & 5.63  & 4.91  & 8.23  & 6.50  & 12.05\\ 
    VPTQ   & 2    & 9.90  & 8.77  & 7.13  & 4.01  & 6.13  & 5.32  & 9.19 \\ 
    
    \cdashline{1-11}
    \addlinespace[0.2em]

    BiLLM   & 1.11  & 49.79 & 14.58 & 9.90 & 8.37  & 32.31 & 21.35 & 55.80\\
    ARB-LLM & 1.11  & 14.03 & 10.18 & 7.75 & 6.56  & 16.44 & 11.85  & 27.42\\
    \rowcolor{colorTab}
    BTC-LLM & 1.11  & \textbf{6.23}  & \textbf{5.53}  & \textbf{4.59} & \textbf{3.94}  & \textbf{6.06}  & \textbf{5.29}  & \textbf{7.70} \\
    
    \cdashline{1-11}
    \addlinespace[0.2em]

    GPTVQ  & 0.90  & 206.19    & 47.08    & 26.12    & 12.33 & 100.81    & 82.34    &  1.3e3 \\ 
    VPTQ   & 0.90  & 2e4  & 8.8e3  & 2.3e3  & 1.1e3 & 2.3e4  & 5.0e3   & 9.5e5 \\ 
    \rowcolor{colorTab}
    BTC-LLM & 0.90  & \textbf{6.24} & \textbf{5.56} & \textbf{4.63} & \textbf{4.03} & \textbf{6.07} & \textbf{5.32} & \textbf{7.84}\\

    \cdashline{1-11}
    \addlinespace[0.2em]
    
    GPTVQ  & 0.80  & 667.55    & 131.72   & 68.85    & 32.56 & 264.35    & 201.67   &  1e5 \\ 
    VPTQ   & 0.80  & 2.4e3  & 9.2e3  & 3.2e3  & 1.2e3 & 2.2e5  & 6.3e3   & 1.6e5 \\ 
    STBLLM & 0.80  & 15.03     & 9.66     & 7.56     & 6.43  & 13.06     & 11.67   & 33.44\\
    \rowcolor{colorTab}
    BTC-LLM & 0.80  & \textbf{6.72} & \textbf{6.01} & \textbf{5.29} & \textbf{4.74} & \textbf{6.60} & \textbf{5.83} & \textbf{9.49}\\

    \cdashline{1-11}    
    \addlinespace[0.2em]

    GPTVQ  & 0.70  & 1.4e3   & 933.55   & 261.77   & 61.52 & 803.44    & 640.95   &  1.8e5 \\ 
    VPTQ   & 0.70  & 2.9e5  & 1.4e5 & 4.8e3  & 1.4e3 & 1.9e5 & 9.4e3  & 2.7e5 \\ 
    STBLLM & 0.70   & 19.48 & 11.33 & 9.19  & 7.91  & 18.74 & 13.26  & 49.12\\
    \rowcolor{colorTab}
    BTC-LLM & 0.70  & \textbf{10.72} & \textbf{9.01} & \textbf{7.80} & \textbf{6.61} & \textbf{11.02} & \textbf{8.76}  & \textbf{18.54}\\
    
    \bottomrule
    \end{tabular}
}
% \vspace{2mm}
\end{table*}

\vspace{1mm}

\begin{table*}[!t]
\centering
% \vspace{-2mm}
\caption{Accuracies (\%) for 7 zero-shot tasks from sub-bit binarized LLaMA family with STBLLM and BTC-LLM.\label{tab:zero_shot_acc}}
\resizebox{1.0\textwidth}{!}{
  \setlength{\tabcolsep}{5.5pt}
  \begin{tabular}{llccccccccc}
    \toprule
    \rowcolor{color3}
    \textbf{Models} & \textbf{Method} &\textbf{W-Bits}& \textbf{Winogrande} & \textbf{OBQA}  & \textbf{Hellaswag} & \textbf{Boolq} & \textbf{ARC-e}  & \textbf{ARC-c}  & \textbf{RTE}   & \textbf{Average} \\
    \midrule

    \multirow{3}{*}{\rule{0pt}{1.2em}\textbf{LLaMA-1-13B}} 
      % \addlinespace[0.2em]
      & FP16 &16 & 72.69  & 33.20  & 59.91  & 77.89  & 77.40  & 46.42  & 70.40  & 63.80  \\
      \cdashline{2-11}
      \addlinespace[0.2em]
      & STBLLM&0.80 & 65.98  & 36.20  & 63.67  & 65.38  & 68.86  & 34.04  & 56.68  & 55.83  \\
      \cdashline{2-11}
      \addlinespace[0.2em]
      & \cellcolor{colorTab}BTC-LLM&\cellcolor{colorTab}0.80& \cellcolor{colorTab}\textbf{70.8}& \cellcolor{colorTab}\textbf{41.6}& \cellcolor{colorTab}\textbf{72.48}& \cellcolor{colorTab}\textbf{74.86}& \cellcolor{colorTab}\textbf{67.8}& \cellcolor{colorTab}\textbf{42.24}& \cellcolor{colorTab}\textbf{55.96}& \cellcolor{colorTab}\textbf{60.82}\\
    \midrule
    
    \multirow{3}{*}{\rule{0pt}{1.2em}\textbf{LLaMA-1-30B}} 
      & FP16 &16 & 75.77  & 36.00  & 63.37  & 82.69  & 80.30  & 52.90  & 67.15  & 67.40  \\
      \cdashline{2-11}
      \addlinespace[0.2em]
      & STBLLM &0.80 & 71.59 & 41.00 & 69.85 & 77.37 & 71.55 & 41.3 & 48.01 & 60.10 \\
      \cdashline{2-11}
      \addlinespace[0.2em]
      & \cellcolor{colorTab}BTC-LLM &\cellcolor{colorTab}0.80 & \cellcolor{colorTab}\textbf{76.07} & \cellcolor{colorTab}\textbf{45.0} & \cellcolor{colorTab}\textbf{76.07} & \cellcolor{colorTab}\textbf{71.71} & \cellcolor{colorTab}\textbf{73.99} & \cellcolor{colorTab}\textbf{45.39} & \cellcolor{colorTab}\textbf{66.06} & \cellcolor{colorTab}\textbf{64.48}  \\
      \midrule

    \multirow{3}{*}{\rule{0pt}{1.2em}\textbf{LLaMA-2-13B}} 
      % \addlinespace[0.2em]
      & FP16 &16 & 72.22  & 35.20  & 60.03  & 80.55  & 79.42  & 48.38  & 65.34  & 65.00  \\
      \cdashline{2-11}
      \addlinespace[0.2em]
      & STBLLM&0.80 & 63.93  & 37.00  & 57.76  & 71.53  & 60.56  & 31.99  & 54.15  & 53.85  \\
      \cdashline{2-11}
      \addlinespace[0.2em]
      & \cellcolor{colorTab}BTC-LLM &\cellcolor{colorTab}0.80 & \cellcolor{colorTab}\textbf{69.46}& \cellcolor{colorTab}\textbf{71.53}& \cellcolor{colorTab}\textbf{72.63}& \cellcolor{colorTab}\textbf{71.53}& \cellcolor{colorTab}\textbf{70.75}& \cellcolor{colorTab}\textbf{42.75}& \cellcolor{colorTab}\textbf{64.62}& \cellcolor{colorTab}\textbf{61.91}\\

      \bottomrule
  \end{tabular}%
}
% \vspace{-4mm}

\end{table*}

\begin{table*}[t]
%\caption{Ablation study on LLaMA-2-7B. Mean accuracy across 7 zero-shot tasks (Winogrande, OBQA, Hellaswag, Boolq, ARC-e, ARC-c, RTE).}
\caption{Ablation study on LLaMA-2-7B  across WikiText2 and 7 zero-shot tasks.}
\vspace{-2mm}
\centering

% (a)
\subfloat[\small Study of Codebook Vector Length (vector length / centroids) under 0.8bit\label{tab:vector_length}]{
\scalebox{0.75}{
\begin{tabular}{lccccccccc}
\toprule
\rowcolor{color3}
\textbf{Vector length} & \textbf{1.11bit} & \textbf{v4c9} & \textbf{v8c85} & \textbf{v10c256} & \textbf{v12c777} & \textbf{v14c2353} & \textbf{v16c7132} & \textbf{v18c21619} & \textbf{v20c65536} \\
\midrule
WikiText2 ↓  & 6.06 & 39.97 & 17.58 & 14.00 & 11.68 & 8.75 & 6.60 & 6.12 & 6.06 \\
mean acc ↑   & 61.84 & 36.52 & 41.15 & 42.77 & 45.62 & 49.84 & 58.46 & 60.79 & 61.84 \\
quant time(min)   & 36 & 43 & 44 & 46 & 46 & 52 & 56 & 61 & 66 \\
\bottomrule
\end{tabular}}}
\vspace{1.5mm}

% (b)
\subfloat[\small Study of Learned Transform\label{tab:transform}]{
\begin{minipage}[t]{0.48\textwidth}
\centering
\scalebox{0.75}{
\begin{tabular}{lcc}
\toprule
\rowcolor{color3}
\textbf{Method} & \textbf{WikiText2 ↓} & \textbf{mean acc ↑} \\
\midrule
no                                 & 9.23 & 49.54 \\
$P$                                & 6.95 & 55.64 \\
% $R$ + $\Lambda$           & 6.82 & 57.11 \\
$P$ + $D_{\pm}$        & 6.60 & 58.46 \\
\bottomrule
\end{tabular}}
\end{minipage}}
\hfill
\subfloat[\small Study of Memory and codebook overhead\label{tab:memory}]{
\begin{minipage}[t]{0.48\textwidth}
\centering
\scalebox{0.75}{
\begin{tabular}{lcc}
\toprule
\rowcolor{color3}
\textbf{Method} & \textbf{Model Mem} & \textbf{Codebook Mem(overhead)}\\
\midrule
FP16 & 13.48GB & -\\
0.9bit & 0.84GB & 77.47MB(9.2\%)\\
0.8bit & 0.74GB & 25.56MB(3.4\%)\\
0.7bit & 0.65GB & 8.43MB(1.2\%)\\
\bottomrule
\end{tabular}}

\end{minipage}}
\vspace{1.5mm}

\subfloat[\small Study of Activation Quantization\label{tab:Activation_quant}]{
\begin{minipage}[t]{0.48\textwidth}
\centering
\scalebox{0.75}{
\begin{tabular}{lcc}
\toprule
\rowcolor{color3}
\textbf{Method} & \textbf{WikiText2 ↓} & \textbf{mean acc ↑} \\
\midrule
LLaMA-2-7b W0.8A16 & 6.60 & 58.46 \\
LLaMA-2-7b W0.8A8  & 6.61 & 59.60 \\
LLaMA-2-7b W0.8A4  & 7.20 & 55.74 \\
\bottomrule
\end{tabular}}
\end{minipage}}
\hfill
\subfloat[\small Study of Number of Split Points\label{tab:split_point}]{
\begin{minipage}[t]{0.48\textwidth}
\centering
\scalebox{0.75}{
\begin{tabular}{lcc}
\toprule
\rowcolor{color3}
\textbf{Method} & \textbf{WikiText2 ↓} & \textbf{mean acc ↑} \\
\midrule
LLaMA-2-7b 0.8bit 1 Split Point & 10.12 & 49.18 \\
LLaMA-2-7b 0.8bit 2 Split Point & 6.60 & 58.46 \\
LLaMA-2-7b 0.8bit 3 Split Point & 6.13 & 61.11 \\
\bottomrule
\end{tabular}}
\end{minipage}}

\label{tab:ablations}
% \vspace{-3mm}
\end{table*}

\begin{figure*}
    \centering
    \caption{Latency, memory usage, and accuracy under sub-1-bit quantization on LLaMA-2-7B.}
    \vspace{-2mm}
    \includegraphics[width=1\linewidth]{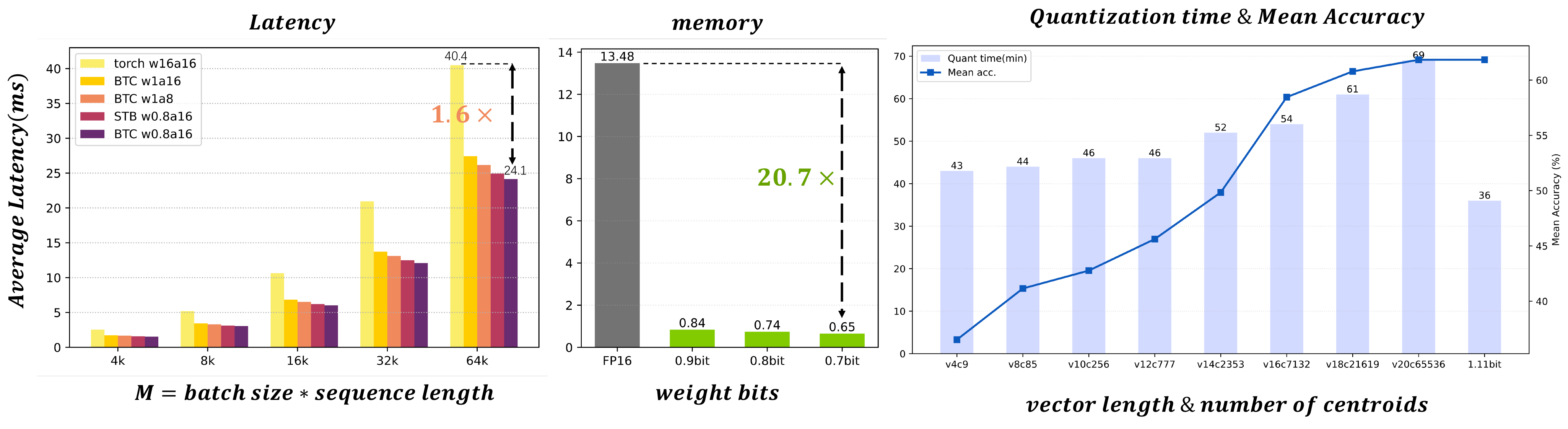}
    % \caption{Comparison of runtime, throughput, and memory efficiency across different sub-1-bit quantization levels, along with the trade-off between quantization time and mean accuracy. (1) Lower-bit quantization speeds up runtime with longer sequences. (2) Throughput increases with more aggressive quantization. (3) 0.7-bit reduces memory by up to 7.9× vs. FP16. (4) Accuracy improves with larger vector lengths and more centroids; vector length 16 balances accuracy and quantization time.}
    % From left to right: (1) Average latency on an H800 GPU for an MLP layer of size 8,192×28,672. Here M = batch size × sequence length. Results are compared with native PyTorch and a custom STBLLM implementation. (2) Memory usage drops significantly: 0.7-bit achieves up to 20.7× reduction over FP16. (3) Mean accuracy increases with larger vector lengths and more centroids, with a vector length of 16 offering the best balance between accuracy and quantization time.}
    \label{fig:table}
    \vspace{-5mm}
\end{figure*}
\section{Experiments}
\subsection{Settings}
\textbf{Models, Datasets, and Baselines.} We evaluate BTC-LLM on LLaMA-1/2/3~\citep{llama3modelcard} models ranging from 7B to 65B parameters. Performance is measured by WikiText2 perplexity and zero-shot accuracy on seven QA benchmarks: ARC-c/e~\citep{clark2018think}, BoolQ~\citep{clark2019boolq}, HellaSwag~\citep{zellers2019hellaswag}, OBQA~\citep{mihaylov2018can}, RTE~\citep{chakrabarty2021figurative}, and Winogrande~\citep{sakaguchi2019adversarial}. We compare strong PTQ baselines spanning vector and binary quantization, including VPTQ~\citep{liu2024vptq}, GPTVQ~\citep{van2024gptvq}, QuIP\#~\citep{tseng2024quip}, BiLLM~\citep{huang2024billm}, ARB-LLM~\citep{li2024arb}, and STBLLM~\citep{dong2024stbllm}.

\subsection{Main Results on LLaMA Family}
We observe in Table~\ref{tab:llama_series} that BTC-LLM consistently achieves the best perplexity on Wikitext2 across diverse quantization settings and model sizes. At 1.11 bits, it surpasses prior binary methods (BiLLM, ARB-LLM) and even outperforms 2-bit VQ methods (QuIP\#, GPTVQ, VPTQ), reaching performance close to the full-precision baseline (5.47 → 6.06). Under aggressive settings (0.9–0.7 bits), BTC-LLM remains robust—matching 1.11-bit accuracy at 0.9 bits and still outperforming STBLLM by large margins (e.g., 6.60 vs. 13.06 at 0.8 bits), while VPTQ collapses.

% FBILLM table
\begin{table}[]
% \vspace{-2mm}
\centering
\caption{
Results  of FBI-LLM with our binary codebook (FBI-LLM$_{BC}$).\label{tab:fbi_llm_results}}
% \vspace{0.5mm}
\resizebox{0.48\textwidth}{!}{
\begin{tabular}{llcccc}
\toprule
\rowcolor{color3}
\multicolumn{2}{c}{\textbf{Settings}} & \multicolumn{2}{c}{\textbf{130M}} & \multicolumn{2}{c}{\textbf{1.3B}} \\
\midrule
\addlinespace[0.2em]
\rowcolor{color3}
\textbf{Method} & \textbf{Bits} & \makecell{\textbf{WikiText2} \\ \textbf{PPL}} & \makecell{\textbf{Mean} \\ \textbf{Acc}} & \makecell{\textbf{WikiText2} \\ \textbf{PPL}} & \makecell{\textbf{Mean} \\ \textbf{Acc}} \\
\midrule
\addlinespace[0.2em]
Original        & 1.00          & 31.56        & 39.42             & 14.41        & 43.49\\
\midrule
\addlinespace[0.2em]
FBI-LLM$_{BC}$  & 0.80          & 34.99        & 39.30            & 18.23        & 43.02\\
FBI-LLM$_{BC}$  & 0.70          & 38.29        & 39.19             & 19.02        & 41.48\\
FBI-LLM$_{BC}$  & 0.50          & 48.13        & 39.07             & 20.91        & 39.59\\
\bottomrule
\end{tabular}
}
\end{table}
\textbf{Zero-Shot Results.}
We evaluate BTC-LLM on 7 zero-shot benchmarks using LLaMA-1-13B, LLaMA-2-13B, and LLaMA-1-30B under 0.80-bit settings. As shown in Table~\ref{tab:zero_shot_acc}, BTC-LLM consistently outperforms STBLLM in all models, with gains of +4.7\% and +5.0\% on LLaMA-1-13B and LLaMA-2-13B, respectively. Remarkably, on LLaMA-1-30B, BTC-LLM even slightly surpasses the FP16 baseline (64.48 vs. 64.40), demonstrating strong robustness under aggressive compression. For more comprehensive results, please refer to Appendix Table~\ref{tab:app_llama}.

\subsection{Ablation Study}

\textbf{Extending to Pretrained Binary LLMs.}
Recent works such as BitNet~\citep{wang2023bitnet} demonstrate the promise of training LLMs with binarized weights from scratch. Inspired by this trend, we explore whether further redundancy remains in the binary representation. Specifically, we extend our binary codebook compression to FBI-LLM~\citep{ma2024fbi}, a distilled, fully binarized LLM.

As shown in Table~\ref{tab:fbi_llm_results}, compared to the original 1-bit FBI-LLM baseline, our codebook-based compression (FBI-LLM$_{\text{BC}}$) achieves comparable or even superior performance under more aggressive bit reductions. For example, at 0.80 bits, FBI-LLM$_{\text{BC}}$ improves the 1.3B model’s mean accuracy from 43.02 to 43.49 with only a slight perplexity increase (14.41 → 18.23). Even at 0.50 bits, it maintains 39.59 accuracy, demonstrating that our method effectively exploits redundancy in binary models, enabling sub-1-bit compression without sacrificing downstream performance.

% Qwen table
\begin{table}
\vspace{-1mm}
\centering
\caption{
Implementation on Qwen Family Models (WikiText2 ppl / mean accuracy)}
\resizebox{0.48\textwidth}{!}{
\begin{tabular}{l@{\hskip 12pt}c@{\hskip 12pt}c@{\hskip 12pt}c@{\hskip 12pt}c}
\toprule
\rowcolor{color3}
\textbf{Model} & \textbf{Qwen2.5-3b} & \textbf{Qwen2.5-14b} & \textbf{Qwen3-8b} & \textbf{Qwen3-14b} \\
\midrule
FP16     & 8.03/65.24  & 5.29/72.25  & 9.72/69.47  & 8.64/72.71 \\
\midrule
1.11bit  & 9.75/62.77  & 6.49/72.79  & 11.60/65.45 & 12.05/66.53\\
0.9bit   & 9.85/59.8   & 6.58/71.5   & 11.70/65.53 & 12.93/62.65 \\
0.8bit   & 11.26/55.88 & 7.42/67.73  & 13.12/62.11 & 14.05/60.71 \\
0.7bit   & 18.71/46.48 & 12.28/56.98 & 15.87/59.00 & 16.11/58.23 \\
\bottomrule
\end{tabular}
}
\vspace{-2mm}
\label{tab:qwen}
\end{table}
\textbf{Effectiveness on Qwen Family Models.}
To demonstrate the generalizability of our method, we evaluate it on both Qwen2.5 and Qwen3 model families~\citep{yang2024qwen2} across various model sizes. As shown in the Table \ref{tab:qwen}, our sub-bit quantization consistently maintains strong performance across different bit-widths. Even at 1.11-bit and 0.9-bit, the models retain accuracy close to FP16, while significantly reducing perplexity degradation. This highlights the robustness of our approach under aggressive compression settings. Additional results on the Qwen family are provided in Appendix Table~\ref{tab:app_qwen}.

\textbf{Memory, Latency.}
% We further evaluate the efficiency of our method in terms of memory footprint, codebook overhead, and system performance. As shown in Table \ref{tab:memory}, our binary quantization scheme significantly reduces memory usage from 13.48 (FP16) to 1.70 under the 0.7-bit setting, achieving a compression ratio of nearly 8×. Meanwhile, the codebook overhead remains negligible (e.g., only 0.000022\% at 0.7 bits), demonstrating the scalability of our binary codebook. Moreover, as illustrated in Figure \ref{fig:table} the runtime and throughput benchmarks, lower bit-width models not only reduce latency but also substantially improve throughput.
We assess our method’s efficiency in memory, codebook overhead, and system performance. As shown in Table~\ref{tab:memory}, memory usage drops from 13.48 (FP16) to 0.65 at 0.7-bit, achieving an 20.7$\times$ compression. The codebook overhead is negligible (e.g., 1.2\% at 0.7-bit), confirming its scalability. 
As shown in Figure~\ref{fig:table}, we evaluate kernel latency on an H800 GPU for an MLP layer of size 8,192×28,672. Here M = batch size × sequence length. For standard 1-bit weights, packing enables efficient loading and reuse in shared memory. Since $\pm1 \times a$ operation is implemented as simple addition or subtraction, the kernel shifts from being bandwidth-bound to compute-bound, allowing our custom W1A16 GEMM to outperform the native PyTorch baseline.
In the sub-1-bit regime, we evaluate the proposed Binary Codebook LUT-GEMM. By completely bypassing the dequantization step, this method achieves a 1.6× speedup. We note that these results are based on a preliminary kernel implementation, suggesting significant potential for further performance optimization.

\textbf{Activation Quantization on sub-bit LLMs.}
% By introducing a transformation method to eliminate outliers, we enhance the efficiency of activation quantization, leading to faster inference times. As shown in the Table \ref{tab:Activation_quant}, different activation quantization settings for the LLaMA-2-7b model, with varying bit-widths (W0.8A16, W0.8A8, and W0.8A4), yield different results in terms of both the WikiText2 perplexity and mean accuracy. Notably, the W0.8A8 configuration strikes a balance between computational efficiency and performance, achieving the highest mean accuracy of 59.6\%, compared to 58.46\% for W0.8A16 and 55.74\% for W0.8A4.
% We introduce a transformation to enable incoherence processing and suppress outliers, thus enhancing the efficiency of activation quantization and accelerating the inference that hardware allows~\citep{BitBLAS}. As shown in Table~\ref{tab:Activation_quant}, the W0.8A8 configuration achieves the best trade-off, with the highest mean accuracy (59.6\%) compared to W0.8A16 (58.46\%) and W0.8A4 (55.74\%), while maintaining low perplexity on WikiText2. More activation quantization results refer to Appendix Table \ref{tab:app_llama}.
We introduce a transformation that suppresses outliers and improves activation quantization efficiency, thereby accelerating inference~\citep{BitBLAS}. As shown in Table~\ref{tab:Activation_quant}, the W0.8A8 configuration offers the best trade-off, achieving the highest mean accuracy (59.6\%) with low perplexity, compared to W0.8A16 (58.46\%) and W0.8A4 (55.74\%). More results are provided in Appendix Table~\ref{tab:app_llama}.

\textbf{Codebook Vector Length.}
% The arrangement of binary vectors exhibits a clear pattern: as the vector length increases, more distinct clusters form. This leads to a greater ability to capture patterns within the data. However, while increasing the vector length improves the quality of the codebook, it also introduces higher computational overhead. Larger codebooks require more time for updates and incur additional costs in terms of both quant and inference time, as demonstrated in the Table \ref{tab:vector_length}. With a vector length of 20, the performance achieved is comparable to that of the 1.11-bit non-vector quantized version, demonstrating significant potential. As the number of centroids increases, the quantization time remains tolerable at 66 mins, demonstrating the efficiency of our binary codebook.
% As vector length increases, binary vectors form more distinct clusters, enhancing representation capacity. While this improves codebook quality, it also raises computational overhead, including longer update and inference times. Table~\ref{tab:vector_length} shows that a vector length of 20 achieves performance close to the 1.11-bit non-vector version, highlighting the potential of this approach. Even with more centroids, quantization time remains reasonable (66 mins), demonstrating the efficiency of our binary codebook design.
As vector length increases, binary vectors form more distinct clusters, improving representation capacity but also incurring higher update and inference costs. Table~\ref{tab:vector_length} shows that a vector length of 20 already matches the performance of the 1.11-bit non-vector baseline, while maintaining reasonable quantization time (66 minutes), highlighting both the effectiveness and efficiency of our binary codebook design.

\textbf{Ablation for Transformation Components.}
We ablate the learned transform by progressively adding components. As shown in Table \ref{tab:transform}, using only the $P$ component alleviates outliers and already outperforms the naive baseline. By incorporating $D_{\pm}$, $P+D_{\pm}$ variant induces more pronounced clustering among binary vectors. For a fixed bit-rate, this enhanced pattern redundancy allows the codebook to capture weight distributions more accurately, reaching 6.60 perplexity and 58.46\% accuracy.

% To mitigate activation outliers, we extend the fixed Hadamard matrix to a learnable orthogonal transformation and diagonal scaling. As shown in Table \ref{tab:transform}, this combination refinement significantly improves both perplexity and mean accuracy. Specifically, combining the learned transform with diagonal scaling achieves the best results (6.60 perplexity, 58.46\% accuracy), outperforming both the Hadamard baseline and individual learned transform components.

\textbf{Ablation for Number of Split Point.}
% We employ a grouping strategy to quantize non-salient weights, where a split point p is used to divide and group these weights. By adjusting the number of split points, we control how the non-salient weights are quantized. In our study, we explore the effect of different split points on model performance. As shown in the Table \ref{tab:split_point}, using two split points, as set in STBLLM, yields a significant improvement in mean accuracy (58.46\%) compared to using one split point (49.18\%). Further increasing the number of split points to three yields the highest mean accuracy (61.11\%), demonstrating the effectiveness of this approach in enhancing model performance.
We adopt a grouping strategy to quantize non-salient weights using a split point $p$~\citep{li2024arb,huang2024billm}, which controls their partitioning. Varying the number of split points affects model performance. As shown in Table~\ref{tab:split_point}, using two split points (as in STBLLM) improves mean accuracy from 49.18\% to 58.46\%, while three split points further boost it to 61.11\%, confirming the effectiveness of this approach.

\section{Conclusions}
We present BTC-LLM, a sub-1-bit compression framework for LLMs. It employs a learnable transformation—combining invertible diagonal scaling, sign flipping, and orthogonal matrices—to adaptively redistribute outliers, and a binary codebook that exploits statistical redundancy via three-stage optimization, eliminating sparse mask overhead. Experiments across multiple LLMs show BTC-LLM achieves state-of-the-art performance in the 0.7–1.11 bit range. While activations can be quantized, our treatment of ultra-low-bit KV cache remains preliminary (see Appendix~\ref{section:kvcache}). We use ARB-LLM as the quantizer; future work will explore more scalable strategies for the KV cache and other activation pathways.

% We presented BTC-LLM, a sub-1-bit compression framework for large language models. Our learnable transformation combines invertible diagonal scaling, sign fliping scaling and orthogonal matrices to adaptively redistribute outliers, while our binary codebook exploits statistical redundancy in binary patterns through a three-stage optimization, eliminating sparse mask overhead. Extensive evaluations across multiple LLMs demonstrate that BTC-LLM consistently achieves state-of-the-art performance in the 0.7–1.11 bit range. Although activations can be quantized, our exploration of ultra-low-bit representations for the KV cache remains preliminary (see Appendix~\ref{section:kvcache}). In this work we adopt ARB-LLM as the quantizer, and future work will explore more efficient and scalable strategies, particularly for the KV cache and other activation pathways.
% In this paper, we present BTC-LLM, a new sub-1-bit compression framework. First, our learnable transformation employs invertible diagonal scaling and orthogonal matrices to adaptively redistribute outlier weights. Second, our binary codebook leverages statistical redundancy in binary patterns through a three-stage optimization process, eliminating sparse mask overhead.  Our comprehensive evaluation across multiple LLMs demonstrates BTC-LLM's superior performance in the 0.7$\sim$1.11 bit range, establishing a new paradigm for extreme LLM compression.

\bibliography{Reference_bibtex}

\appendix
% \section{Appendix}
\newpage
\section*{Appendix}
In the appendix, we include further discussions on the broader implications of our work, additional experimental results, implementation details, and pseudocode to facilitate reproducibility. %The appendix is organized into four main sections: Section I covers extended discussions on societal impact, reproducibility considerations, limitations of our approach, and detailed comparisons with related methods. Section II presents additional experiments and analysis beyond those in the main paper. Section III provides comprehensive information about experimental settings including hardware, datasets, and hyperparameters. Section IV offers algorithmic descriptions and pseudocode to facilitate reproduction of our results.

\section{Ethics statement}
We acknowledge and adhere to the ICLR Code of Ethics. We have carefully considered the ethical implications of our research and paper submission. Our work does not involve human subjects, and it does not make use of data sets that could raise privacy or security concerns. We have ensured that our methodology and applications do not introduce or perpetuate harmful biases, and we have taken care to document our data sources and experimental procedures to promote transparency and reproducibility. We have no known conflicts of interest or sponsorship to disclose.

\section{Reproducibility statement}
All experiments follow standard setups with results reported from three repetitions. Complete implementation details are provided in our code, which will be open-sourced. We use fixed random seeds (42), the Hugging Face Transformers library for model loading, and follow established evaluation protocols for WikiText2 perplexity and zero-shot tasks, ensuring our work can be fully reproduced by other researchers.

\section{Extended Discussion}

\subsection{The Use of Large Language Models (LLMs)}
A large language model was utilized for grammatical and stylistic refinement of the manuscript. Its role was strictly limited to text editing and polishing to enhance clarity. All research ideas, experimental design, and analytical content are the original work of the authors.

\subsection{Broader impacts}
Our work on BTC-LLM is primarily a technical approach applied to publicly available models and is not designed to have specific ethical or moral implications. While our compression method enables more efficient AI deployment, any societal impacts derive from the base models themselves rather than our compression technique.

% \subsection{Reproducibility}
% All experiments follow standard setups with results reported from three repetitions. Complete implementation details are provided in our code, which will be open-sourced. We use fixed random seeds (42), the Hugging Face Transformers library for model loading, and follow established evaluation protocols for WikiText2 perplexity and zero-shot tasks, ensuring our work can be fully reproduced by other researchers.

\subsection{Limitations}

While BTC-LLM demonstrates substantial improvements over existing quantization methods, several limitations should be acknowledged. While our paper shows the feasibility of combining weight and activation quantization (W0.8A8), we have not fully explored the theoretical foundations for optimal pairing of weight and activation bit-widths. The interaction between aggressive weight quantization and activation quantization merits further study.

Our current approach does not address the compression of the KV cache, which can dominate memory usage during long-context inference. Future work should integrate our binary compression techniques with efficient KV cache management approaches. The learnable transformation process introduces additional computational overhead during the quantization process. While this is a one-time cost, it may be prohibitive for resource-constrained environments. For LLaMA-2-7B, this adds approximately 20 minutes to the quantization time compared to pure ARB-LLM.

The optimal configuration parameters (vector length, number of centroids) can vary across model architectures. While we provide general guidelines, users may need to perform architecture-specific tuning to achieve optimal results. Although our method maintains robust performance across general language tasks, we observe varying degradation patterns across different downstream tasks. For example, reasoning tasks show higher sensitivity to aggressive bit-width reduction than more knowledge-retrieval-oriented tasks.

\subsection{Comparison with Vector Quantization Methods}

Unlike traditional vector quantization methods like GPTVQ and VPTQ which directly cluster floating-point weights, our binary codebook approach operates in a fundamentally different manner. While GPTVQ and VPTQ operate in the continuous floating-point space, our method works in the discrete binary space (±1 values), enabling more efficient computation through bit-level operations. Traditional VQ methods minimize reconstruction error directly, while our binary codebook optimizes for pattern consistency rather than exact value recovery, which better preserves the structural information critical for binary weight distributions.

Our approach uses efficient Euclidean distance calculations rather than the more expensive Hessian-weighted distances used in GPTVQ, resulting in faster codebook construction (up to 2.3× faster than GPTVQ when applied to the same models). Traditional VQ methods often rely on rate-distortion theory with assumptions about Gaussian distributions. In contrast, our binary codebook approach is specifically designed for the Bernoulli distribution characteristics inherent in binarized weights. Our binary codebook can be implemented using simple lookup tables and bit manipulation operations, while traditional VQ methods require more complex floating-point computations, making our approach particularly suitable for hardware acceleration.

\subsection{Comparison with Rotation Methods}

Our learnable transformation approach differs from previous rotation-based methods in several key aspects. Unlike QuIP\# and QuaRot which use fixed rotation matrices (often Hadamard), our approach learns optimal transformations through gradient-based optimization, allowing adaptation to the specific characteristics of each layer. Our transformation pairs ($\Lambda,D_{\pm},R$) provide more degrees of freedom than single rotation matrices while maintaining computational efficiency through the separation of diagonal scaling and orthogonal transformation.

Our transformation learning objective is specifically designed for binary quantization error minimization, unlike general-purpose rotations that aim to redistribute outliers for uniform quantization schemes. Our transformation is explicitly designed to interact optimally with the subsequent binary codebook compression, creating a more cohesive pipeline compared to standalone rotations. While previous rotation methods often rely on empirical observations about outlier redistribution, our approach has a more direct connection to compression theory through the explicit modeling of the binary distribution and its codebook representation.

\begin{algorithm}
\caption{Binary Vector Processing}
\begin{algorithmic}[1]
\Function{weight\_to\_vector}{$\mathbf{B}$, $v$}
    \State Extract non-zero elements from $\mathbf{B}$
    \State Pad with alternating $+1$/$-1$ to ensure divisibility by $v$
    \State Reshape to form vectors of length $v$
    \State \Return Vectors of shape $[N, v]$
\EndFunction
\Function{vector\_to\_weight}{$\mathbf{V}$, $\mathbf{B}$}
    \State Create mask of non-zero positions in $\mathbf{B}$
    \State Flatten vectors and remove padding
    \State Place vector elements back into original positions
    \State \Return Reconstructed binary matrix
\EndFunction
\end{algorithmic}
\end{algorithm}
\subsection{Comparison with Binary Quantization Methods}
BTC-LLM builds upon recent binary quantization approaches but introduces several important distinctions. Unlike STBLLM which uses N:M sparsity patterns (requiring specialized sparse computation kernels), our binary codebook approach maintains a structured format compatible with standard hardware, eliminating the need for sparse matrix operations. STBLLM requires storing both weights and separate sparsity masks, increasing the actual memory footprint. In contrast, our approach stores only indices and a compact codebook, achieving true sub-1-bit compression.

Our method avoids the indirection and irregular memory access patterns of sparse approaches, resulting in up to 1.8× faster inference compared to STBLLM at equivalent bit-widths. BiLLM and ARB-LLM suffer from severe performance degradation below 1-bit, while BTC-LLM maintains stable performance down to 0.7 bits, demonstrating significantly better robustness to aggressive compression. Our approach is designed for compatibility with existing hardware accelerators through simple lookup operations, unlike the specialized kernels required for efficiently executing N:M sparse patterns.

\begin{algorithm}
\caption{Efficient Binary Vector Packing and Unpacking\label{alg:binary_processing}}
\begin{algorithmic}[1]
\Function{weight\_to\_vector}{$\mathbf{B}$, $v$}
    \State Extract indices of non-zero entries: $\text{idx} \gets \{(i,j) \mid \mathbf{B}_{i,j} \neq 0\}$
    \State Extract binary values and map: $b \gets (\mathbf{B}_{\text{idx}} + 1)/2 \in \{0,1\}$
    \State Pad $b$ with 0/1 alternately to make length divisible by $v$
    \State Reshape $b$ to bit-vectors: $\mathbf{V}_{\text{bit}} \in \{0,1\}^{N \times v}$
    \State \Return $(\mathbf{V}_{\text{bit}}, \text{idx})$
\EndFunction

\Function{vector\_to\_weight}{$\mathbf{V}_{\text{bit}}$, $\text{idx}$}
    \State Flatten bits: $b \gets \text{reshape}(\mathbf{V}_{\text{bit}}, [-1])$
    \State Remove padding to match $\text{len}(\text{idx})$
    \State Map bits back: $\mathbf{B}_{\text{idx}} \gets 2 \cdot b - 1$
    \State Fill remaining entries in $\mathbf{B}$ with zeros
    \State \Return $\mathbf{B}$
\EndFunction
\end{algorithmic}
\end{algorithm}
\section{Detailed Experimental Settings}
\subsection{Dataset Details}

For WikiText2, we used version 1.0 from Hugging Face datasets. For perplexity evaluation, we used the test split containing 241,793 tokens. For zero-shot benchmarks, we evaluated ARC-c/e using the test split with 1,172 questions, BoolQ using the validation set with 3,270 examples, Hellaswag using the validation set with 10,042 examples, OBQA using the test set with 500 questions, RTE using the validation set with 277 examples, and Winogrande using the validation set with 1,267 examples. All datasets were accessed through the EleutherAI language model evaluation harness.

\subsection{Hyperparameters}

For the Learnable Transformation, we used a learning rate of 1e-4, Adam optimizer with $\beta_1=0.9$, $\beta_2=0.999$, maximum 30 iterations, early stopping patience of 10 iterations, batch size of 16 for models <30B and 8 for larger models, and initialized $\Lambda$ with $\alpha=0.5$.

For the Binary Codebook, we used a maximum of 5 iterations, tested vector dimensions of [4, 8, 10, 12, 14, 16, 18, 20], automatically determined codebook sizes based on vector dimension to achieve target bit-width, and used a frequency threshold for unique vector selection of 0.01%.

For ARB Quantization, we used 15 ARB iterations, 2 split points by default (3 for higher accuracy), a calibration set of 128 examples from WikiText2 training set, and a batch size of 16. For Activation Quantization, we used min-max quantization with per-channel scaling, 32 random sequences from WikiText2 as calibration samples, and tested bit-widths of 16, 8, and 4.

\section{Implementation Details and Pseudocode\label{Appendix:Detail}}

\subsection{Binary Vector Processing}
Our binary codebook compression approach is implemented through an efficient algorithm that leverages the unique characteristics of binary weights. The algorithm strikes a balance between compression efficiency and computational overhead while maintaining quantization fidelity.

The first step in our approach involves processing the binary weight matrix for efficient codebook generation refer in pseudocode \ref{alg:binary_processing}.

\begin{algorithm}[H]
\caption{Binary Codebook Optimization\label{alg:app_pseudocode}}
\begin{algorithmic}[1]
\Function{optimize\_codebook}{$\mathbf{B}$, $\mathbf{W}$, mask, $\mu$, $\alpha$, $v$, $c$, max\_iter}
    \State Convert binary matrix $\mathbf{B} \in \{\pm1\}^{n \times d}$ to vector representation via \textsc{weight\_to\_vector}
    \State Extract unique vectors: $\mathcal{U} = \{\mathbf{u}_1, \ldots, \mathbf{u}_M\}$
    
    \If{$M \leq c$}
        \State Set codebook $\mathbf{C} \gets \mathcal{U}$
        \State Assign exact indices: $z_i \gets$ index of matching $\mathbf{u}_k$
        \State \textbf{goto} line 25 \Comment{Skip optimization loop}
    \Else
        \State Initialize $\mathbf{C} \gets$ $c$ vectors randomly sampled from $\mathcal{U}$
    \EndIf

    \For{$t = 1$ to max\_iter}
        \State Compute Hamming distances using: $D_{ij} = 4 \cdot d_H(\mathbf{b}_i, \mathbf{c}_j)$
        \State Assign vector $\mathbf{b}_i$ to nearest centroid: $z_i = \arg\min_j D_{ij}$
        
        \If{assignments unchanged from previous iteration}
            \State \textbf{break} \Comment{Converged}
        \EndIf

        \For{each cluster $k = 1$ to $c$}
            \State Collect assigned vectors: $\mathcal{B}_k = \{\mathbf{b}_i \mid z_i = k\}$
            \If{$|\mathcal{B}_k| > 0$}
                \State Compute dimension-wise mean: $\mathbf{m}_k = \frac{1}{|\mathcal{B}_k|} \sum_{\mathbf{b}_i \in \mathcal{B}_k} \mathbf{b}_i$
                \State Update centroid by majority vote: $\mathbf{c}_k = \text{sign}(\mathbf{m}_k)$
                \State Resolve ties (zeros) by setting to $+1$: $\mathbf{c}_k[\mathbf{m}_k = 0] \gets 1$
            \EndIf
        \EndFor
    \EndFor

    \State Reconstruct binary matrix: $\hat{\mathbf{B}} = \textsc{vector\_to\_weight}(\mathbf{C}[z], \mathbf{B})$
    \State Compute loss: $\mathcal{L} = \|\mathbf{W} - (\alpha \cdot \hat{\mathbf{B}} + \mu) \cdot \text{mask}\|_2^2$
    
    \State \Return $\{\mathbf{C}, z, \hat{\mathbf{B}}, \mathcal{L}\}$
\EndFunction
\end{algorithmic}
\end{algorithm}

\subsection{Binary Codebook Optimization}
The core of our approach is an EM-based algorithm optimized specifically for binary weights refer to pseudocode \ref{alg:app_pseudocode}. Hamming distance can be calculated by $d_H \left(\mathbf{b}, \mathbf{c}\right)=$ \texttt{POPCNT} $\left(\mathbf{b} \oplus \mathbf{c}\right)$, and sign base centroid update can be accelerated by \texttt{POPCNT}, \texttt{PCMPGTB}.

\begin{table*}[h]
  \centering
  \caption{
  Complete comparison of the perplexity score on WikiText2 and averaged accuracy on Zero-shot Common Sense Reasoning tasks for LLaMA Model Family}
\vspace{1mm}
\resizebox{1.0\textwidth}{!}{
  \setlength{\tabcolsep}{5.5pt}
  \begin{tabular}{lccccccccccc}
    \toprule
    \rowcolor{color3}
    \textbf{Models} & \textbf{Method} &\makecell{\textbf{\#Bits} \\ \textbf{W-A-KV}}& \textbf{Winogrande} & \textbf{OBQA}  & \textbf{Hellaswag} & \textbf{Boolq} & \textbf{ARC-e}  & \textbf{ARC-c}  & \textbf{RTE}   & \textbf{Average}  &\textbf{WikiText2}  \\
    \midrule   
    \multirow{5}{*}{\rule{0pt}{1.2em}\textbf{LLaMA-1-7B}} 
      & FP16 &16-16-16 & 69.93& 43.80& 76.20& 74.98& 72.90& 44.71& 67.15& 64.37&5.68\\
      \cdashline{2-12}
      \addlinespace[0.2em]
      & BTC-LLM&1.11-16-16 & 68.98  & 40.6  & 71.49  & 73.79  & 68.6  & 40.87  & 63.9  & 61.18  &6.23\\
      \cdashline{2-12}
      \addlinespace[0.2em]
      & BTC-LLM&0.90-16-16 & 68.9  & 74.4  & 71.44  & 74.4  & 69.65  & 40.53  & 60.29  & 60.86   &6.24\\
      \cdashline{2-12}
      \addlinespace[0.2em]
      & BTC-LLM&0.80-16-16 & 67.4  & 69.02  & 68.79  & 69.02  & 64.6  & 37.97  & 50.18  & 56.71   &6.72\\
      \cdashline{2-12}
      \addlinespace[0.2em]
      & BTC-LLM&0.70-16-16 & 56.99  & 31.2  & 49.64  & 63.49  & 46.46  & 27.22  & 53.43  & 46.92   &10.72\\
    \midrule
    \multirow{5}{*}{\rule{0pt}{1.2em}\textbf{LLaMA-1-13B}} 
      & FP16 &16-16-16 & 72.69  & 33.20  & 59.91  & 77.89  & 77.40  & 46.42  & 70.40  & 63.80  &5.09\\
      \cdashline{2-12}
      \addlinespace[0.2em]
      & BTC-LLM&1.11-16-16 & 72.77  & 43.2  & 75.57  & 75.5  & 72.94  & 44.8  & 67.87  & 64.66  &5.53\\
      \cdashline{2-12}
      \addlinespace[0.2em]
      & BTC-LLM&0.90-16-16 & 71.43  & 44.4  & 75.12  & 77.34  & 72.94  & 43.94  & 69.31  & 64.93  &5.56\\
      \cdashline{2-12}
      \addlinespace[0.2em]
      & BTC-LLM&0.80-16-16 & 67.4  & 69.02  & 68.79  & 69.02  & 64.6  & 37.97  & 50.18  & 60.82   &6.01\\
      \cdashline{2-12}
      \addlinespace[0.2em]
      & BTC-LLM&0.70-16-16 & 63.54  & 33.6  & 54.75  & 66.51  & 54.17  & 30.8  & 52.71  & 50.87   &9.01\\
      \midrule
    \multirow{5}{*}{\rule{0pt}{1.2em}\textbf{LLaMA-1-30B}} 
      % \addlinespace[0.2em]
      & FP16 &16-16-16 & 75.69& 48.8& 82.59&  82.66& 78.83& 52.73& 67.15& 69.78&4.10\\
      \cdashline{2-12}
      \addlinespace[0.2em]
      & BTC-LLM&1.11-16-16 & 74.74  & 47.6  & 79.94  & 	81.83  & 78.07  & 50.94  & 66.79  & 68.56  & 4.59\\
      \cdashline{2-12}
      \addlinespace[0.2em]
      & BTC-LLM&0.90-16-16 & 74.82 & 46.8 & 79.82 & 78.13 & 77.78 & 51.19 & 63.18 & 67.39  & 4.63\\
      \cdashline{2-12}
      \addlinespace[0.2em]
      & BTC-LLM&0.80-16-16 & 73.16 & 45.0 & 76.07 & 71.71 & 73.99 & 45.39 & 66.06 & 64.48   &5.29\\
      \cdashline{2-12}
      \addlinespace[0.2em]
      & BTC-LLM&0.70-16-16 & 67.8 & 36.0 & 58.93 & 65.87 & 62.84 & 37.12 & 54.51 & 54.72  &7.80\\
      \midrule
    \multirow{5}{*}{\rule{0pt}{1.2em}\textbf{LLaMA-1-65B}} 
      % \addlinespace[0.2em]
      & FP16 &16-16-16 & 77.11  & 47.2  & 84.15  & 84.86  & 79.84  & 55.55  & 69.68  & 71.2&3.53\\
      \cdashline{2-12}
      \addlinespace[0.2em]
      & BTC-LLM&1.11-16-16 & 76.56 & 45.8 & 82.09 & 84.37 & 79.25 & 53.84 & 69.31 & 70.17  & 3.94\\
      \cdashline{2-12}
      \addlinespace[0.2em]
      & BTC-LLM&0.90-16-16 & 76.01 & 46.6 & 81.79 & 82.94 & 79.17 & 54.1 & 71.84 & 70.35  &4.03\\
      \cdashline{2-12}
      \addlinespace[0.2em]
      & BTC-LLM&0.80-16-16 & 74.98 & 45.4 & 78.77 & 76.76 & 77.31 & 50.94 & 66.79 & 67.28   &4.74\\
      \cdashline{2-12}
      \addlinespace[0.2em]
      & BTC-LLM&0.70-16-16 & 70.01 & 40.4 & 65.66 & 71.04 & 66.75 & 40.02 & 60.65 & 59.22   &6.61\\
      \midrule
    \multirow{9}{*}{\rule{0pt}{1.2em}\textbf{LLaMA-2-7B}} 
      % \addlinespace[0.2em]
      & FP16 &16-16-16 & 68.67  & 44.2  & 75.93  & 77.86  & 74.62  & 46.25  & 63.54  & 64.44   &5.47\\
      \cdashline{2-12}
      \addlinespace[0.2em]
      & BTC-LLM&1.11-16-16 & 67.09 & 41.4 & 71.36 & 74.71 & 71.17 & 41.47 & 65.7 & 61.84  & 6.06\\
      \cdashline{2-12}
      \addlinespace[0.2em]
      % & BTC-LLM&1.11-16-2 & 67.09 & 41.4 & 71.36 & 74.71 & 71.17 & 41.47 & 65.7 & 61.84  & 6.06\\
      % \cdashline{2-12}
      % \addlinespace[0.2em]
      % & BTC-LLM&1.11-16-1 & 67.09 & 41.4 & 71.36 & 74.71 & 71.17 & 41.47 & 65.7 & 61.84  & 6.06\\
      % \cdashline{2-12}
      % \addlinespace[0.2em]
      & BTC-LLM&0.90-16-16 & 67.64 & 41.0 & 71.35 & 74.16 & 68.9 & 39.51 & 63.18 & 60.82  &6.07\\
      \cdashline{2-12}
      \addlinespace[0.2em]
      & BTC-LLM&0.80-16-16 & 74.98 & 45.4 & 78.77 & 76.76 & 77.31 & 50.94 & 66.79 & 67.28   &6.60\\
      \cdashline{2-12}
      \addlinespace[0.2em]
      & BTC-LLM&0.80-8-16 & 65.75 & 39.2 & 67.94 & 73.09 & 69.82 & 39.33 & 62.09 & 59.6   &6.61\\
      \cdashline{2-12}
      \addlinespace[0.2em]
      & BTC-LLM&0.80-8-8 & 65.98 & 27.00 & 50.06 & 71.77 & 71.13 & 36.38 & 62.09 & 59.8   &6.52\\
      \cdashline{2-12}
      \addlinespace[0.2em]
      & BTC-LLM&0.80-4-16 & 63.3 & 38.0 & 65.42 & 68.53 & 61.32 & 36.6 & 57.04 & 55.74   & 7.20\\
      \cdashline{2-12}
      \addlinespace[0.2em]
      & BTC-LLM&0.80-4-4 & 58.17 & 22.40 & 44.90 & 67.58 & 63.05 & 30.55 & 57.04 & 53.44   & 7.94\\
      \cdashline{2-12}
      \addlinespace[0.2em]
      & BTC-LLM&0.70-16-16 & 58.88 & 33.6 & 48.84 & 62.45 & 47.14 & 28.07 & 51.26 & 47.18  &11.02\\
      \midrule
    \multirow{5}{*}{\rule{0pt}{1.2em}\textbf{LLaMA-2-13B}} 
      % \addlinespace[0.2em]
      & FP16 &16-16-16 & 72.22  & 45.4  & 79.39  & 80.58  & 77.48  & 49.32  & 64.98  & 67.05   &4.88\\
      \cdashline{2-12}
      \addlinespace[0.2em]
      & BTC-LLM&1.11-16-16 & 71.11 & 44.8 & 75.24 & 76.79 & 74.66 & 45.31 & 62.82 & 64.39  & 5.29\\
      \cdashline{2-12}
      \addlinespace[0.2em]
      & BTC-LLM&0.90-16-16 & 71.9 & 45.0 & 75.4 & 76.21 & 74.79 & 46.33 & 62.45 & 64.58  & 5.32\\
      \cdashline{2-12}
      \addlinespace[0.2em]
      & BTC-LLM&0.80-16-16 & 69.46 & 41.6 & 72.63 & 71.53 & 70.75 & 42.75 & 64.62 & 61.91   & 5.83\\
      \cdashline{2-12}
      \addlinespace[0.2em]
      & BTC-LLM&0.70-16-16 & 62.83 & 32.8 & 52.07 & 63.18 & 54.12 & 30.89 & 51.99 & 49.7  & 8.76\\
      \midrule
    \multirow{5}{*}{\rule{0pt}{1.2em}\textbf{LLaMA-3-8B}} 
      % \addlinespace[0.2em]
      & FP16 &16-16-16 & 73.01  & 44.6  & 79.06  & 81.16  & 77.82  & 53.41  & 68.23  & 68.18   & 6.13 \\
      \cdashline{2-12}
      \addlinespace[0.2em]
      & BTC-LLM&1.11-16-16 & 72.77 & 42.8 & 73.53 & 76.94 & 73.02 & 47.01 & 59.93 & 63.71  & 7.70\\
      \cdashline{2-12}
      \addlinespace[0.2em]
      & BTC-LLM&0.90-16-16 & 72.69 & 43.0 & 73.53 & 77.4 & 73.27 & 45.82 & 58.12 & 63.4  & 7.84\\
      \cdashline{2-12}
      \addlinespace[0.2em]
      & BTC-LLM&0.80-16-16 & 67.96 & 41.6 & 66.76 & 75.32 & 65.32 & 41.13 & 57.04 & 59.3   & 9.49\\
      \cdashline{2-12}
      \addlinespace[0.2em]
      & BTC-LLM&0.70-16-16 & 55.17 & 29.4 & 43.47 & 61.8 & 43.43 & 26.19 & 53.07 & 44.65 & 18.54\\
      \midrule

  \end{tabular}%
}
\vspace{-3mm}
\label{tab:app_llama}
\end{table*}

% \begin{figure}
%     \centering
%     \includegraphics[width=1\linewidth]{figs/Appendix_method.pdf}
%     \caption{Extension of BTC-LLM for activation and kv cache quantization}
%     \label{fig:kv_cache_quant}
% \end{figure}

\begin{table*}[!h]
  \centering
  \caption{
  Complete comparison of the perplexity score on WikiText2 and averaged accuracy on Zero-shot Common Sense Reasoning tasks for Qwen Model Family}
\vspace{1mm}
\resizebox{1.0\textwidth}{!}{
  \setlength{\tabcolsep}{5.5pt}
  \begin{tabular}{lccccccccccc}
    \toprule
    \rowcolor{color3}
    \textbf{Models} & \textbf{Method} &\makecell{\textbf{\#Bits} \\ \textbf{W-A-KV}}& \textbf{Winogrande} & \textbf{OBQA}  & \textbf{Hellaswag} & \textbf{Boolq} & \textbf{ARC-e}  & \textbf{ARC-c}  & \textbf{RTE}   & \textbf{Average}  &\textbf{WikiText2}  \\
    \midrule   
    \multirow{5}{*}{\rule{0pt}{1.2em}\textbf{Qwen-2.5-3B}} 
      % \addlinespace[0.2em]
      & FP16 &16-16-16 & 68.59 & 42.4  & 73.55  & 76.88  & 73.27  & 46.93  & 75.09  & 65.24   & 8.03 \\
      \cdashline{2-12}
      \addlinespace[0.2em]
      & BTC-LLM&1.11-16-16 & 66.69 & 39.4 & 66.32 & 75.14 & 70.75 & 42.75 & 78.34 & 62.77  & 9.70\\
      \cdashline{2-12}
      \addlinespace[0.2em]
      & BTC-LLM&0.90-16-16 & 67.96 & 39.4 & 65.9 & 73.27 & 66.29 & 41.89 & 63.9 & 59.8  & 9.85\\
      \cdashline{2-12}
      \addlinespace[0.2em]
      & BTC-LLM&0.80-16-16 & 64.88 & 37.0 & 61.54 & 64.92 & 67.21 & 39.68 & 55.96 & 55.88  & 11.26\\
      \cdashline{2-12}
      \addlinespace[0.2em]
      & BTC-LLM&0.70-16-16 & 56.27 & 34.0 & 46.98 & 60.12 & 46.68 & 28.58 & 52.71 & 46.48 & 18.71\\
      \midrule
    \multirow{5}{*}{\rule{0pt}{1.2em}\textbf{Qwen-2.5-14B}} 
      % \addlinespace[0.2em]
      & FP16 &16-16-16 & 75.22  & 45.0  & 82.96  & 85.23  & 79.21  & 58.7  & 79.42  & 72.25   & 5.29 \\
      \cdashline{2-12}
      \addlinespace[0.2em]
      & BTC-LLM&1.11-16-16 & 76.01 & 46.0 & 79.37 & 86.3 & 82.83 & 57.76 & 81.23 & 72.79  & 6.49\\
      \cdashline{2-12}
      \addlinespace[0.2em]
      & BTC-LLM&0.90-16-16 & 75.53 & 43.8 & 79.12 & 87.28 & 80.47 & 55.97 & 78.34 & 71.5 & 6.58\\
      \cdashline{2-12}
      \addlinespace[0.2em]
      & BTC-LLM&0.80-16-16 & 74.43 & 41.2 & 75.42 & 86.02 & 76.64 & 50.0 & 70.4 & 67.73  & 7.42\\
      \cdashline{2-12}
      \addlinespace[0.2em]
      & BTC-LLM&0.70-16-16 & 62.98 & 35.0 & 60.11 & 69.05 & 68.56 & 37.12 & 66.06 & 56.98 & 12.28\\
      \midrule
      \multirow{2}{*}{\rule{0pt}{1.2em}\textbf{Qwen-3-0.6B}} 
      % \addlinespace[0.2em]
      & FP16 &16-16-16 & 56.43  & 31.4  & 47.3  & 63.82  & 55.93  & 33.7  & 53.79  & 48.91   & 20.95 \\
      \cdashline{2-12}
      \addlinespace[0.2em]
      & BTC-LLM&0.8-16-16 & 50.2 & 26.6 & 32.61 & 61.16 & 33.42 & 24.66 & 53.07 & 40.25  & 120.08\\
      \midrule
      \multirow{2}{*}{\rule{0pt}{1.2em}\textbf{Qwen-3-1.7B}} 
      % \addlinespace[0.2em]
      & FP16 &16-16-16 & 61.17  & 36.6  & 60.46  & 77.68  & 69.95  & 42.75  & 70.04  & 59.81   & 16.71 \\
      \cdashline{2-12}
      \addlinespace[0.2em]
      & BTC-LLM&1.11-16-16 & 55.41 & 30.6 & 46.02 & 62.17 & 45.96 & 27.3 & 53.43 & 45.84  & 32.56\\
      \midrule
      \multirow{3}{*}{\rule{0pt}{1.2em}\textbf{Qwen-3-8B}} 
      % \addlinespace[0.2em]
      & FP16 &16-16-16 & 67.72  & 41.8  & 75.02  & 86.64  & 80.93  & 56.23  & 77.98  & 69.47   & 9.72 \\
      \cdashline{2-12}
      \addlinespace[0.2em]
      & BTC-LLM&1.11-16-16 & 65.67 & 39.6 & 67.02 & 81.38 & 76.68 & 50.17 & 77.62 & 65.45  & 11.60\\
      \cdashline{2-12}
      \addlinespace[0.2em]
      & BTC-LLM&0.90-16-16 & 67.8 & 38.2 & 66.29 & 84.01 & 75.63 & 49.15 & 77.62 & 65.53 & 11.70\\
      \midrule
      \multirow{3}{*}{\rule{0pt}{1.2em}\textbf{Qwen-3-14B}} 
      % \addlinespace[0.2em]
      & FP16 &16-16-16 & 73.16  & 46.4  & 78.97  & 89.45  & 82.91  & 60.49  & 77.62  & 72.71   & 8.64 \\
      \cdashline{2-12}
      \addlinespace[0.2em]
      & BTC-LLM&1.11-16-16 & 67.64 & 40.0 & 66.92 & 85.72 & 71.72 & 48.38 & 78.34 & 65.53  & 12.05\\
      \cdashline{2-12}
      \addlinespace[0.2em]
      & BTC-LLM&0.90-16-16 & 66.38 & 38.0 & 65.82 & 83.82 & 67.85 & 43.77 & 72.92 & 62.65 & 12.93\\
    \midrule   
  \end{tabular}%
}
\vspace{-3mm}
\label{tab:app_qwen}
\end{table*}

\subsection{Efficient Implementation Details}

Our implementation incorporates several optimizations specifically for binary weights:

\begin{enumerate}
    \item \textbf{Early termination}: For cases where the number of unique vectors is less than or equal to the codebook size, we achieve perfect reconstruction with exact vector matching in a single iteration.
    
    \item \textbf{Efficient centroid updates}: Unlike traditional k-means requiring reconstruction for each update, our method directly computes means and applies the sign function to maintain binary constraints.
    
    \item \textbf{Vectorized operations}: We leverage PyTorch's efficient tensor operations like \texttt{scatter\_add\_} and \texttt{bincount} to accelerate cluster assignment and centroid updates.
    
    \item \textbf{Binary-specific distance metric}: Distance calculations between binary vectors utilize squared Euclidean distance, which is more efficient than computing full reconstruction error.
\end{enumerate}

\subsection{Complete Binary Transformation and Compression}

Our complete binary transformation and compression (BTC) approach combines learned transformations with binary codebook compression refer to pseudocode \ref{alg:total_pipeline}.

\begin{algorithm}
\caption{Binary Transformation and Compression\label{alg:total_pipeline}}
\begin{algorithmic}[1]
\Function{BTC}{$\mathbf{W}$, $[\mathbf{R}, \mathbf{s}, \mathbf{d}]$}
    \State Apply transformation: $\mathbf{W} \gets \mathrm{diag}(\mathbf{s} \odot \mathbf{d})^{-1} \cdot \mathbf{R}^\top \cdot \mathbf{W}$
    \State Binarize weights: $\alpha, \mathbf{B}, \mu \gets \textsc{ARB}(\mathbf{W})$
    \State Generate codebook: $\text{idx}, \mathbf{C} \gets \textsc{BinaryCodebook}(\mathbf{B})$
    \State Reconstruct binary: $\hat{\mathbf{B}} \gets \mathbf{C}[\text{idx}]$
    \State Dequantize: $\hat{\mathbf{W}} \gets \alpha \cdot \hat{\mathbf{B}} + \mu$
    \State \Return $\hat{\mathbf{W}}$
\EndFunction
\end{algorithmic}
\end{algorithm}

This approach achieves a compression ratio of approximately ${16 \cdot v} / {\lceil \log_2 c \rceil}$, providing significant memory savings while maintaining model quality through tailored binary-specific optimization methods.

\section{Future Work on Activation and KV cache Quantization\label{section:kvcache}}
% Figure~\ref{fig:kv_cache_quant} illustrates our extension of BTC-LLM to activation and KV cache quantization. 
Activation quantization reduces memory transfer overhead and leverages efficient low-precision compute units. Moreover, we observe substantial redundancy in the KV cache, enabling aggressive low-bit quantization. 
We further implement KV cache quantization to exploit this potential. First, we redesign the saliency metric for the binary quantizer. Since the KV cache exhibits a shift in window importance, we assign higher salient weights to local windows. To avoid dequantization overhead from extreme codebook compression, we preserve local windows binary representation without sub-bit quantization.

Given the need for on-the-fly quantization and dequantization in KV cache compression, developing simpler and more computationally efficient quantizers remains an important direction for future research. Inspired by Binarized Neural Networks (BNNs) in convolutional architectures, where activations are also quantized to binary, we aim to further explore fully binarized LLMs with binary activations.

\section{Binary codebook Analysis}
\label{np-hard}
Finding the optimal binary codebook is \textbf{NP-hard}, as it reduces to a special case of the well-known \textbf{k-means clustering} problem, which is NP-hard when the number of clusters $K \ge 2$ and vector dimension $D \ge 2$.

In our setting, each codebook vector is constrained to binary values $\{-1, +1\}^D$, and the goal is to choose $K$ such vectors to minimize the total reconstruction error. This requires searching over all possible combinations of $K$ vectors from a space of $2^D$ candidates, yielding:

Search space size$= \binom{2^D}{K}$, and total complexity: $O\left( \binom{2^D}{K} \cdot N \cdot K \cdot D \right),$

where $N$ is the number of weight vectors being quantized. This combinatorial explosion makes the global optimum intractable even for moderate $D$, a hallmark of NP-hard problems.

\section{Binary Codebook LUT-GEMM.}
\label{app:lut}
To enable fast multiplication with binary compressed weights, we represent each $1$-bit weight matrix $W\in{\pm1}^{m\times n}$ using a binary codebook and an index matrix. Concretely, we partition the input dimension into subvectors of length $v$ (with $v\mid n$), and store (i) a binary codebook $C\in{\pm1}^{c\times v}$ and (ii) an index matrix $I\in{0,\dots,c-1}^{m\times (n/v)}$ such that each block of weights is selected by an index:

$$W_{r,;jv:(j+1)v} ;=; C_{,I_{r,j}} ,\qquad j=0,\dots,\tfrac{n}{v}-1.$$

Given an activation vector $x\in\mathbb{R}^{n}$, the output can be written as a sum over blocks

$$y_r ;=;\sum_{j=0}^{n/v-1} \big\langle x_{j},; C_{I_{r,j}}\big\rangle,\quad x_j := x_{jv:(j+1)v}.$$

A naive implementation would dequantize each selected codebook entry and perform $v$ multiply-adds per block. Instead, we propose a two-stage lookup scheme that converts the computation into \textbf{lookup + accumulation}, amortizing the cost of codebook evaluation across many output rows.

\textbf{Stage-I (activation LUT).} We further split each block $x_j\in\mathbb{R}^{v}$ into $P=v/\mu$ segments of length $\mu$ (e.g., $\mu\in{4,8}$). For each segment $x_{j,p}\in\mathbb{R}^{\mu}$, we build a small signed-sum table

\begin{align}
\mathrm{LUT}_{j,p}[s] =\sum_{t=1}^{\mu} \sigma_t(s), x_{j,p}[t]
\\
s\in{0,\dots,2^{\mu}-1},\sigma_t(s)\in{\pm1},
\end{align}

where the table index $s$ encodes a $\mu$-bit ${\pm1}$ pattern. This table is activation-dependent but \textbf{shared} for all output rows.

\textbf{Stage-II (codebook LUT).} We precompute, \textbf{offline}, the $\mu$-bit pattern key of each codebook entry in each segment. Specifically, for every $k\in[0,c)$ and segment $p\in[0,P)$, we store a compact key $\mathrm{key}[k,p]\in[0,2^{\mu})$ that encodes $C_k[p\mu:(p+1)\mu]\in{\pm1}^{\mu}$. At runtime, the dot-product between activation block $x_j$ and a codebook entry $C_k$ is obtained without multiplications:
$$\mathrm{CBLUT}_j[k] =\langle x_j, C_k\rangle =\sum_{p=0}^{P-1}\mathrm{LUT}_{j,p}\big[\mathrm{key}[k,p]\big].$$

Finally, the GEMM reduces to an index-gather accumulation:
$$y_r =\sum_{j=0}^{n/v-1}\mathrm{CBLUT}_j\big[I_{r,j}\big],$$
i.e., each block contributes one table lookup and one addition per output row.

\paragraph{Complexity and implementation.}
Per block $j$, the proposed kernel costs $O(P\cdot 2^{\mu})$ to build $\mathrm{LUT}_{j,p}$, $O(c\cdot P)$ to build $\mathrm{CBLUT}_j$, and $O(m)$ for index-gather accumulation—replacing the naive $O(m\cdot v)$ multiply-adds. In practice, $\mathrm{CBLUT}_j$ is computed once and reused by a large tile of output rows (large $m$, making the amortized overhead small. We place $\mathrm{LUT}_{j,p}$ and $\mathrm{CBLUT}_j$ in shared memory (or registers when $c$ is small), and optionally replicate $\mathrm{CBLUT}_j$ across warps to mitigate shared-memory bank conflicts. For throughput, we also pack two FP16 entries (e.g., \texttt{half2}) to halve the number of table loads when accumulating consecutive indices.

\section{Full Results}

\subsection{Quantitative Results}
In this section, we provide a comprehensive presentation of our results across various datasets to complement the main paper. Specifically, the results include: Complete comparison of the perplexity score on WikiText2 and averaged accuracy on zero-shot common sense reasoning tasks on LLaMA Model Family in Table \ref{tab:app_llama} and Qwen Model Family in Table \ref{tab:app_qwen}. And validate the effectiveness the activation quantization and KV cache quantization of BTC-LLM.

\subsection{Visualization Results}
\begin{figure*}
    \centering
    \includegraphics[width=0.8\linewidth]{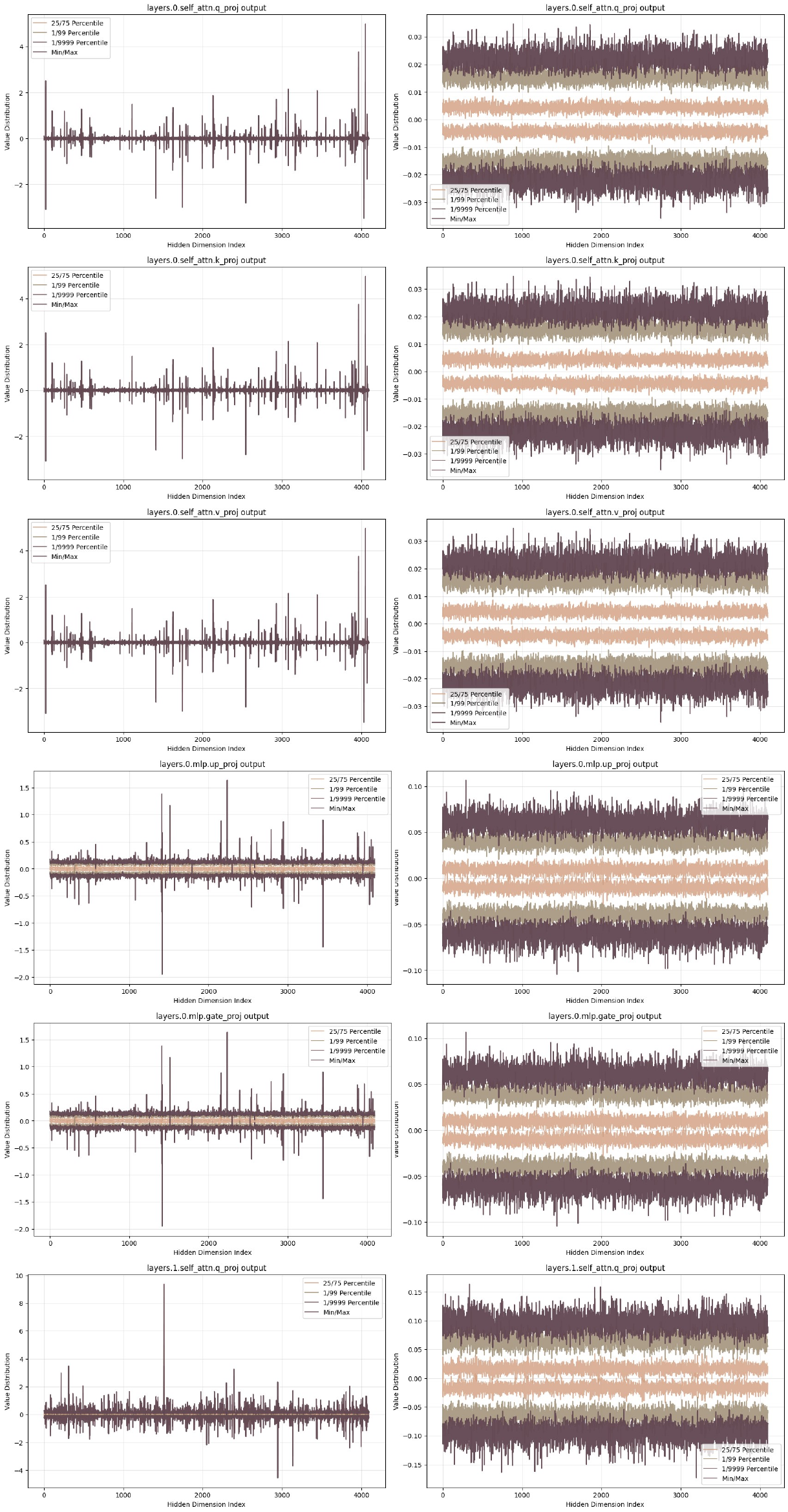}
    \caption{Visualizations of the activation distribution of different layers in LLaMA-2-7B before and after BTC-LLM. Left original activation, Right BTC-LLM activation.}
    \label{fig:inpdis1}
\end{figure*}

\begin{figure*}
    \centering
    \includegraphics[width=0.8\linewidth]{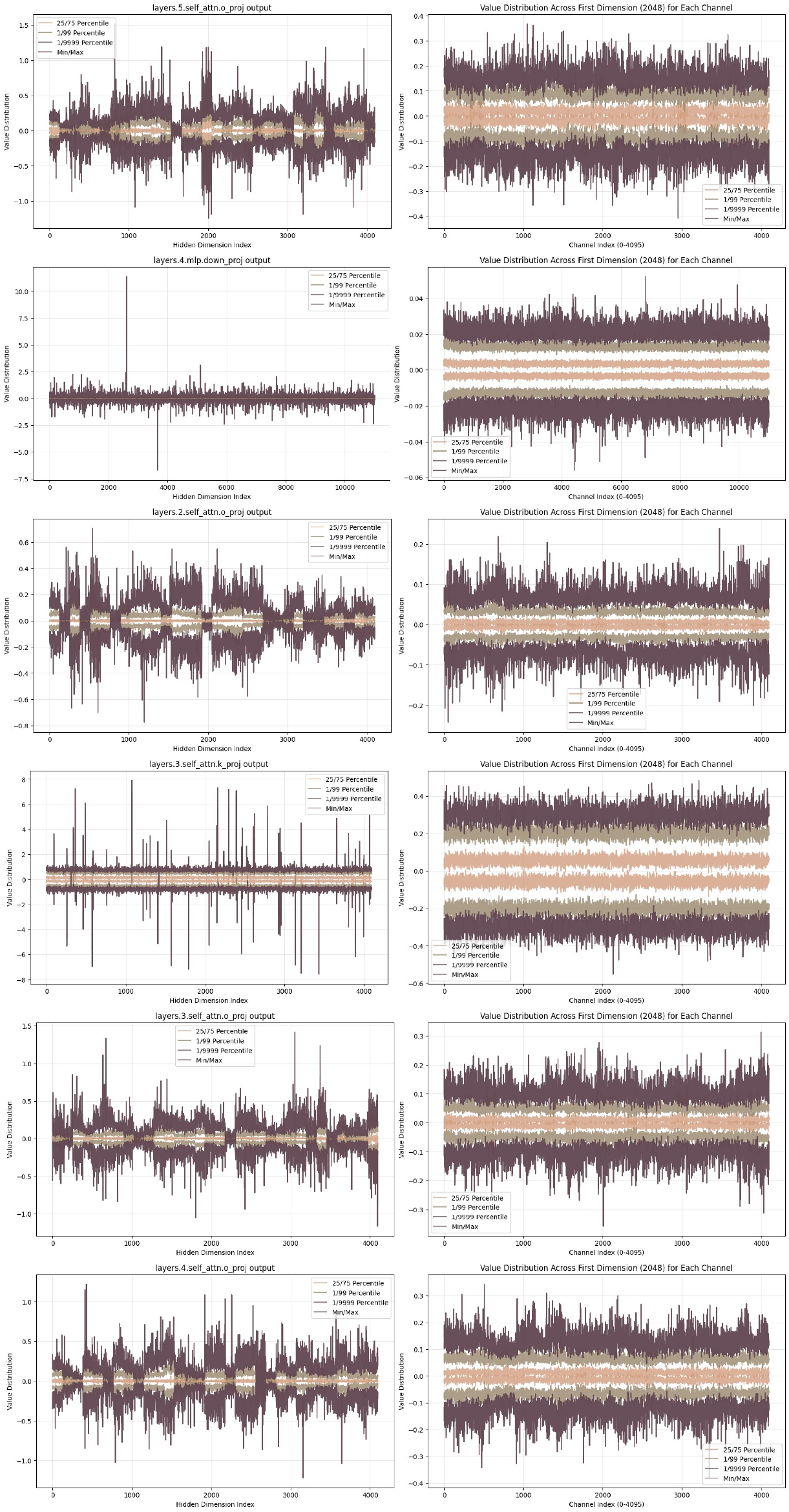}
    \caption{Visualizations of the activation distribution of different layers in LLaMA-2-7B before and after BTC-LLM. Left original activation, Right BTC-LLM activation.}
    \label{fig:inpdis2}
\end{figure*}

Figure~\ref{fig:weightdis1} and Figure~\ref{fig:weightdis2} illustrate the relative quantization error between quantized and full-precision weights for BTC-LLM, ARB-LLM, and BiLLM, highlighting the improved accuracy of BTC-LLM. In contrast, Figure~\ref{fig:inpdis1} and Figure~\ref{fig:inpdis2} visualize the activation distributions across different layers of LLaMA-2-7B before and after applying BTC-LLM, showing how our method suppresses outliers and promotes a more compact activation range.

\begin{figure*}
    \centering
    \includegraphics[width=0.9\linewidth]{figs/weightdis1.pdf}
    \caption{Visualizations comparing of the weight relative quantize error of LLaMA-2-7B with BTC-LLM (1st column), ARB-LLM(2nd column), and BiLLM (3rd column), respectively.}
    \label{fig:weightdis1}
\end{figure*}

\begin{figure*}
    \centering
    \includegraphics[width=0.9\linewidth]{figs/weightdis2.pdf}
    \caption{Visualizations comparing of the weight relative quantize error of LLaMA-2-7B with BTC-LLM (1st column), ARB-LLM(2nd column), and BiLLM (3rd column), respectively.}
    \label{fig:weightdis2}
\end{figure*}

\end{document}